\documentclass[10pt,twocolumn,letterpaper]{article}

\usepackage{cvpr}              
\definecolor{cvprblue}{rgb}{0.21,0.49,0.74}
\usepackage[pagebackref,breaklinks,colorlinks,allcolors=cvprblue]{hyperref}

\usepackage{pifont}
\usepackage{comment}
\usepackage[dvipsnames]{xcolor}
\usepackage{booktabs}
\usepackage{array}
\usepackage{tabularx}
\usepackage{colortbl}
\usepackage{tikz}
\usepackage{multirow}
\usetikzlibrary{arrows.meta, shapes.geometric, shapes,shadows,arrows,backgrounds}
\usepackage{wasysym}

\usepackage{fontawesome5}

\def\paperID{} 
\def\confName{CVPR}
\def\confYear{2026}

\title{SALMUBench: A Benchmark for Sensitive Association-Level Multimodal Unlearning}

\author{
Cai Selvas-Sala$^{1,2,3}$\quad
Lei Kang$^{1,3}$\quad
Lluis Gomez$^{1,3}$\\
$^{1}$Computer Vision Center \quad
$^{2}$Universitat Politècnica de Catalunya\quad
$^{3}$Universitat Autònoma de Barcelona\\
{\tt\small\{cselvas, lkang, lgomez\}@cvc.uab.cat} \\[2pt]
{\small\url{http://cvc-mmu.github.io/salmubench}}
}

\begin{document}
\maketitle
\begin{abstract}
As multimodal models like CLIP become integral to downstream systems, the need to remove sensitive information is critical. However, machine unlearning for contrastively-trained encoders remains underexplored, and existing evaluations fail to diagnose fine-grained, association-level forgetting. We introduce SALMUBench (Sensitive Association-Level Multimodal Unlearning), a benchmark built upon a synthetic dataset of 60K persona-attribute associations and two foundational models: a \textit{Compromised} model polluted with this data, and a \textit{Clean} model without it. To isolate unlearning effects, both are trained from scratch on the same 400M-pair \texttt{retain} base, with the \textit{Compromised} model additionally trained on the \texttt{sensitive} set. We propose a novel evaluation protocol with structured holdout sets (\texttt{holdout\_identity}, \texttt{holdout\_association}) to precisely measure unlearning efficacy and collateral damage. Our benchmark reveals that while utility-efficient deletion is feasible, current methods exhibit distinct failure modes: they either fail to forget effectively or over-generalize by erasing more than intended. SALMUBench sets a new standard for comprehensive unlearning evaluation, and we publicly release our dataset, models, evaluation scripts, and leaderboards to foster future research.
\end{abstract}    
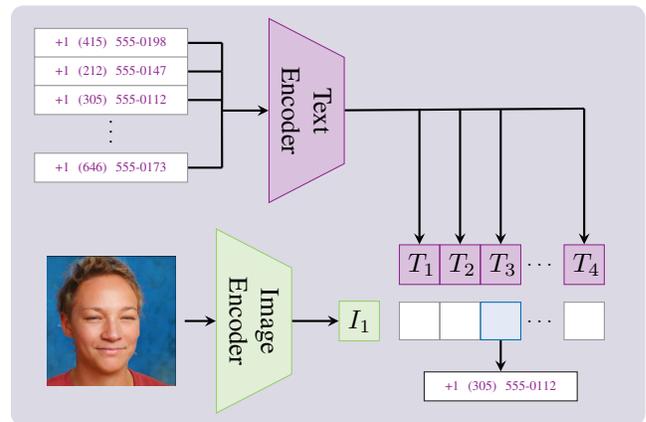
\begin{figure}
    \centering
    \begin{tikzpicture}[background rectangle/.style={draw=CadetBlue!20,fill=CadetBlue!20,rounded corners=1ex}, show background rectangle]

\tikzstyle{ImgEncoder} = [draw=YellowGreen, trapezium, fill=YellowGreen!25, minimum width=10mm, text centered, minimum height=10mm]
\tikzstyle{TxtEncoder} = [draw=Plum, trapezium, fill=Plum!25, text width=15mm, text centered, minimum height=10mm]
\tikzstyle{phone} = [draw=Gray, rectangle, fill=white, text=Plum, text width=13mm, text centered, minimum height=3mm, node distance=6mm]

\tikzstyle{txtfeat} = [draw=Plum, rectangle, fill=Plum!25, text width=3mm, text centered, minimum height=5mm, minimum width=5mm, node distance=5mm]
\tikzstyle{imgfeat} = [draw=YellowGreen, rectangle, fill=YellowGreen!25, text width=3mm, text centered, minimum height=5mm, minimum width=5mm, node distance=5mm]
\tikzstyle{dotprod} = [draw=Gray, rectangle, fill=white, text width=3mm, text centered, minimum height=5mm, minimum width=5mm, node distance=5mm]
\tikzstyle{dotprodtop} = [draw=NavyBlue, rectangle, fill=NavyBlue!10, text width=3mm, text centered, minimum height=5mm, minimum width=5mm, node distance=5mm]
\tikzstyle{line} = [draw, thick]  
\tikzstyle{arrow} = [thick,->,>=stealth]
\tikzstyle{label} = [node distance=5mm]

\node [node distance=0mm] (Image) {\includegraphics[width=1.71cm]{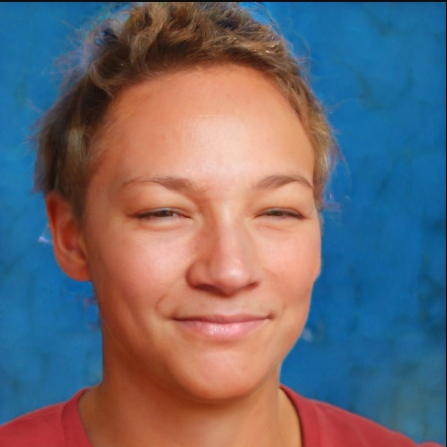}};
\node [ImgEncoder, trapezium stretches body, right of=Image, text width=11.8mm, xshift=9mm, rotate=-90] (ImgEnc) {Image Encoder};
\node[imgfeat, right of=ImgEnc, xshift=9mm](I1){$I_1$};

\node [TxtEncoder, trapezium stretches body, above of=ImgEnc, text width=11.8mm, xshift=7mm, yshift=18mm, rotate=-90] (TxtEnc) {Text Encoder};

\node [phone, above of=Image, xshift=0mm, yshift=31mm, text width=18mm] (phone1) {\tiny+1 (415) 555-0198};
\node [phone, below of=phone1, xshift=0mm, yshift=2.2mm, text width=18mm] (phone2) {\tiny+1 (212) 555-0147};
\node [phone, below of=phone2, xshift=0mm, yshift=2.2mm, text width=18mm] (phone3) {\tiny+1 (305) 555-0112};
\node [label, below of=phone3, xshift=0mm, yshift=0.5mm, rotate=-90] (phonedots) {\scriptsize$\dots$};
\node [phone, below of=phone3, xshift=0mm, yshift=-3mm, text width=18mm] (phone4) {\tiny+1 (646) 555-0173};

\node [label, right of=phone1, xshift=11mm, yshift=0mm] (label1) {};
\node [label, right of=phone2, xshift=11mm, yshift=0mm] (label2) {};
\node [label, right of=phone3, xshift=11mm, yshift=0mm] (label3) {};
\node [label, right of=phone4, xshift=11mm, yshift=0mm] (label4) {};

\node [label, left of=TxtEnc, xshift=-7.5mm, yshift=0mm] (label5) {};

\node[dotprod, right of=I1, xshift=3mm](P1){};
\node[dotprod, right of=P1, xshift=0.4mm](P2){};
\node[dotprodtop, right of=P2, xshift=0.4mm](P3){};
\node[dotprod, right of=P3, xshift=6.1mm](P4){};
\node [label, right of=P3, xshift=0.4mm, yshift=0mm] (Pdots) {\scriptsize$\dots$};

\node[txtfeat, above of=P1, yshift=2.5mm](T1){$T_1$};
\node[txtfeat, above of=P2, yshift=2.5mm](T2){$T_2$};
\node[txtfeat, above of=P3, yshift=2.5mm](T3){$T_3$};
\node[txtfeat, above of=P4, yshift=2.5mm](T4){$T_4$};
\node [label, right of=T3, xshift=0.4mm, yshift=0mm] (Tdots) {\scriptsize$\dots$};

\node [label, above of=T1, yshift=17mm] (auxT1) {};
\node [label, above of=T2, yshift=17mm] (auxT2) {};
\node [label, above of=T3, yshift=17mm] (auxT3) {};
\node [label, above of=T4, yshift=17mm] (auxT4) {};

\node [label, right of=T4, xshift=0.3mm, yshift=0mm] (auxborder1) {};
\node [label, left of=Image, xshift=-5.5mm, yshift=0mm] (auxborder2) {};

\node [phone, draw=Black, below of=P3, xshift=0mm, yshift=-2.7mm, text width=18mm] (phoneP3) {\tiny+1 (305) 555-0112};

\draw [arrow] (auxT1) -- (T1);
\draw [arrow] (auxT2) -- (T2);
\draw [arrow] (auxT3) -- (T3);
\draw [arrow] (auxT4) -- (T4);

\draw [arrow] (P3) -- (phoneP3);

\draw [arrow] (Image) -- (ImgEnc);

\draw [line] (phone1) -- (label1);
\draw [line] (phone2) -- (label2);
\draw [line] (phone3) -- (label3);
\draw [line] (phone4) -- (label4);

\draw [line] (label1.west) -- (label2.west);
\draw [line] (label2.west) -- (label3.west);
\draw [line] (label3.west) -- (label5.east);
\draw [line] (label4.west) -- (label5.east);

\draw [arrow] (label5) -- (TxtEnc);

\draw [arrow] (ImgEnc) -- (I1);
\draw [line] (TxtEnc) -- (auxT4.south);

\end{tikzpicture}
    \caption{CLIP models can memorize and leak private information. CLIP-based systems can associate a face with sensitive attributes (e.g., a phone number) seen during training. 
    }
    \label{fig:teaser}
\end{figure}


\section{Introduction}
The rise of large-scale vision-language models (VLMs) such as CLIP~\cite{OpenAICLIPOriginal} has significantly impacted how visual and textual information is jointly modeled and utilized. Pretrained CLIP embeddings are now foundational across diverse downstream tasks, including image retrieval, text-to-image generation, image captioning, and visual question answering. Yet, as training data for these models is often sourced from vast, uncurated web corpora, these models might inadvertently memorize sensitive information~\cite{carlini2022quantifying} as illustrated in Figure~\ref{fig:teaser}. With increased regulatory attention on data privacy -- by legislation such as the EU General Data Protection Regulation (GDPR) and the associated ``Right to be Forgotten'' -- there is a pressing need to develop effective machine unlearning techniques that allow models to selectively remove learned sensitive information.

While machine unlearning has seen significant progress in the unimodal setting -- with methods proposed for both computer vision~\cite{bourtoule2021machine, golatkar2020eternal, chen2023boundary} and NLP models, including large language models~\cite{jang2023knowledge, guo2020certified} -- the multimodal domain remains comparatively underexplored. Recent efforts such as MultiDelete~\cite{cheng2024multidelete}, CLIPErase~\cite{yang-etal-2025-cliperase}, and zero-shot class unlearning in CLIP~\cite{kravets2025zero} address unlearning in VLMs. However, their unlearning targets do not specifically address private information, and often the compromised model is only exposed to the data to forget via finetuning with small datasets -- e.g., Flickr30K in \cite{cheng2024multidelete} -- limiting practical relevance under the ``Right to be Forgotten'' scenario.

Concurrently, to avoid the ethical and scientific challenges of using real Personally Identifiable Information (PII), the field has increasingly adopted a synthetic-data paradigm. This approach, established in unimodal unlearning by TOFU~\cite{maini2024tofu} and extended to multimodal benchmarks like FIUBench~\cite{FIUBench} and MLLMU-Bench~\cite{liu2024protecting}, enables controlled, privacy-safe evaluation with ground-truth knowledge of what to forget. However, these benchmarks primarily focus on generative MLLMs (multimodal chatbots) and evaluate forgetting via Visual Question Answering (VQA) tasks, leaving a critical gap in evaluating embedding-based foundation models like CLIP. Furthermore, they typically inject target knowledge via fine-tuning, making it difficult to isolate unlearning effects from pretraining artifacts. CLIP models pose distinct challenges: they are contrastively trained from scratch on vast datasets, serve as foundational backbones in larger systems, and encode information in non-generative, embedding-based form.

To bridge this gap, we introduce SALMUBench, a novel benchmark explicitly designed to evaluate machine unlearning methods in CLIP models. Our benchmark consists of a synthetic dataset of 60,000 image-text pairs associating fictitious personas with sensitive private information, such as names, locations, email addresses, phone numbers, and credit card numbers. Crucially, unlike existing benchmarks, we train two large-scale ViT/B-16 CLIP models entirely from scratch using approximately 400 million samples over 32 epochs: a \textit{Compromised} model exposed to the sensitive synthetic data, and a \textit{Clean} model that serves as an ideal reference model. This controlled setup allows measuring unlearning performance by ensuring a model's predictions become identical to those of a model retrained from scratch without the sensitive data~\cite{guo2020certified}, free from the ambiguities introduced by incremental fine-tuning.

Our core contribution is a novel evaluation protocol built on structured holdout sets (\texttt{holdout\_identity} and \texttt{holdout\_association}). This enables, for the first time, nuanced diagnosis of unlearning failures, distinguishing between ineffective, catastrophically damaging, and over-generalizing methods. Our experiments demonstrate that no current method solves this trade-off, highlighting substantial room for improvement. We publicly release our dataset, models, and evaluation tools to foster future research. SALMUBench thus sets a new standard for the comprehensive evaluation of machine unlearning in multimodal embedding-based models.
\section{Related Work}
\label{sec:background}

Multimodal machine unlearning remains largely underexplored compared to unimodal settings. Recent work has begun addressing this gap, notably with MultiDelete and CLIPErase -- two frameworks targeting vision-language models like CLIP. Both aim to forget specific image-text associations while preserving model utility.

MultiDelete~\cite{cheng2024multidelete} introduces a tri-objective loss to (a) decouple paired modalities in the forget set, (b) retain unimodal performance on the remaining data, and (c) preserve multimodal alignment. It achieves this by scrambling image-text pairings in the forget set while maintaining accuracy on retained image-only, text-only, and multimodal predictions. Their experiments show strong forgetting with minimal performance degradation.

CLIPErase~\cite{yang-etal-2025-cliperase} extends this philosophy with three specialized modules: a Forgetting Module that minimizes cross-modal similarity for the forget set, a Retention Module that applies CLIP’s contrastive loss to preserve alignment in the retain set, and a Consistency Module that penalizes distributional drift from the original model. This tri-modular design enables precise removal of undesired associations (e.g., selectively forgetting ``apple'' the fruit but retaining Apple Inc.) while preserving overall performance across retrieval and generation tasks.

In contrast, Kravets \etal \cite{kravets2025zero} focus on unlearning specific classes in CLIP's zero-shot classifier. They generate synthetic prototypes of the target class using gradient ascent, avoiding the need for original real data. Then, by applying Lipschitz regularization to flatten the model's sensitivity to that class, they iteratively fine-tune CLIP until it no longer identifies the concept, updating only the most relevant layers to minimize collateral damage.

\begin{figure*}[ht]
\centering
\scalebox{0.85}{
\begin{tikzpicture}[
  box/.style={draw=CadetBlue, fill=CadetBlue!20, rectangle, align=center, minimum height=2.8cm},
  arr/.style={-{Stealth[length=3mm]}, line width=1.2pt, CadetBlue!70!black}
]
  \node (step0) [box, minimum width=2.9cm, text width=3.1cm] at (0,0)
    {\includegraphics[width=1.5cm]{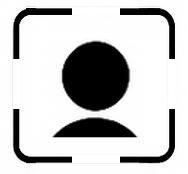}\\[0.3em]
     \textbf{(1) Anchor Seeding}};

  \node (step1) [box, minimum width=3.1cm, text width=3.5cm] at (4.2,0)
    {\includegraphics[width=1.8cm]{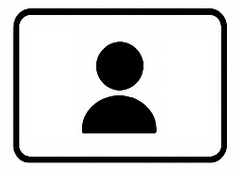}\\[0.3em]
     \textbf{(2) Identity-Preserving Generation}};
     
  \node (step2) [box, minimum width=3.2cm, text width=3.1cm] at (8.4,0)
    {\includegraphics[width=1.8cm]{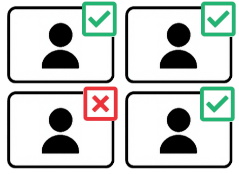}\\[0.3em]
     \textbf{(3) CLIP-based Filtering/Curation}};
     
  \node (step3) [box, minimum width=2.7cm, text width=3.1cm] at (12.4,0)
    {\includegraphics[width=1.8cm]{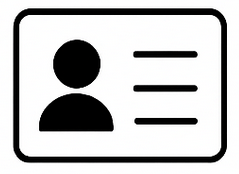}\\[0.3em]
     \textbf{(4) PII Assignment}};
     
  \node (step4) [box, minimum width=2.7cm, text width=3.1cm] at (16.4,0)
    {\includegraphics[width=1.8cm]{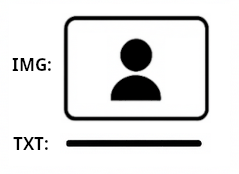}\\[0.3em]
     \textbf{(5) LLM Caption Paraphrasing}};
     
  \draw[arr] (step0) -- (step1);
  \draw[arr] (step1) -- (step2);
  \draw[arr] (step2) -- (step3);
  \draw[arr] (step3) -- (step4);
\end{tikzpicture}
}
\caption{The construction pipeline of the SALMU dataset follows five sequential stages.}
\label{fig:salmu_pipeline}
\end{figure*}
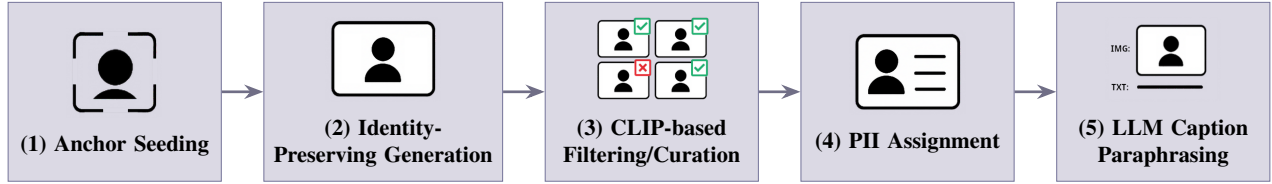

Beyond CLIP, some recent works target multimodal large language models (MLLMs) -- i.e., vision-enhanced chatbots. Benchmarks such as MLLMU-Bench \cite{liu2024protecting} and FIUBench \cite{FIUBench} have been introduced to evaluate machine unlearning techniques in this multimodal setting. Both benchmarks adapt the synthetic-data-driven evaluation framework introduced by TOFU \cite{maini2024tofu} -- a benchmark originally designed for unimodal large language models using a fictitious author dataset -- to the multimodal domain. The use of synthetic data enables these benchmarks to comprehensively evaluate how effectively sensitive/private information can be forgotten in multimodal models under the ``Right to be Forgotten'' scenario, specifically in Visual Question Answering (VQA) tasks.

These existing benchmarks inspire our work, but they differ substantially from our proposed benchmark in several aspects. First, we specifically focus on machine unlearning within contrastively-trained foundation models like CLIP. CLIP's embedding-based architecture fundamentally differs from the generative multimodal language models targeted by existing benchmarks. Instead of evaluating forgetting via QA tasks in generative chat-based models, our benchmark directly evaluates forgetting capabilities within pretrained embedding spaces. This is important, as pretrained CLIP embeddings are widely used as foundational modules across diverse downstream architectures and tasks. Second, while prior benchmarks primarily rely on fine-tuning pretrained models on synthetic datasets to inject the sensitive/private knowledge intended for deletion, our approach provides two large-scale CLIP models pre-trained entirely from scratch. This strategy offers a robust, realistic, and precisely controlled setting for evaluating machine unlearning methods, free from potential confounding factors introduced by incremental fine-tuning. Finally, our benchmark differs significantly from existing evaluations of CLIP unlearning methods \cite{cheng2024multidelete, kravets2025zero}, whose unlearning targets do not specifically address personal or private information, limiting their relevance under the ``Right to be Forgotten''. 

\section{SALMUBench: Design and Components}
\label{sec:design}

The SALMUBench framework provides a complete ecosystem for association-level unlearning evaluation through three integrated components. We detail (1) the construction of our synthetic dataset simulating sensitive associations; (2) the from-scratch pre-training of our foundational \textit{Compromised} and \textit{Clean} models for a controlled setup; and (3) a comprehensive evaluation protocol to assess unlearning efficacy and utility preservation.

\subsection{Synthetic Dataset Construction}
\label{sec:synth_data}

The foundation of our benchmark is the SALMU dataset, a synthetic corpus designed for a comprehensive and reproducible evaluation of text-image association-level machine unlearning. As illustrated in Figure~\ref{fig:salmu_pipeline}, its construction follows five sequential stages: (1) anchor seeding from a face dataset; (2) identity-preserving image generation; (3) CLIP-based filtering and demographic curation; (4) sensitive PII attribute assignment; and (5) LLM-based caption paraphrasing.

The pipeline proceeds through the following five stages (full details, including model identifiers, prompt examples, and filtering rules, are provided in supplementary material):

\paragraph{(1) Anchor Seeding.}
We select 1,000 reference face synthetic images from the SFHQ Part~4 dataset~\cite{SFHQP4} to serve as identity anchors. These images carry no associated metadata and are used solely to seed visual appearance.

\paragraph{(2) Identity-Preserving Generation.}
For each persona, we generate $\sim$100 images using IP-Adapter-FaceID Plus~\cite{IP-Adapter} conditioned on the anchor. Textual prompts are constructed programmatically by combining phrases across semantic categories (camera angle, shot scale, action, expression, setting) to ensure visual diversity. Images are generated at $1024{\times}1024$, then outpainted via ControlNet~\cite{ControlNet} to realistic aspect ratios sampled from the DataComp corpus~\cite{DataComp} distribution, and resized to a maximum of 512\,px.

\paragraph{(3) CLIP-Based Filtering and Curation.}
A pretrained CLIP model~\cite{FilterCLIP, OpenCLIPRepo} assigns each image zero-shot labels for visual distortions, multiple subjects, gender, and ethnicity (FairFace categories~\cite{FairFace}). We discard flagged images, establish a majority-vote demographic profile per persona, remove images inconsistent with that profile, and drop personas with fewer than 50 remaining images, yielding 774 coherent identities. Independent validation with an off-the-shelf face recognition model (InsightFace \texttt{buffalo\_l}) confirms a sharp separation: intra-persona cosine similarity $0.660 {\pm} 0.087$ vs.\ inter-persona $0.101 {\pm} 0.084$; a manual audit of 1,000 random images estimates a noise rate of 0.10\%.

\paragraph{(4) Personally Identifiable Information (PII) Assignment.}
Each persona receives a culturally-consistent profile of sensitive attributes: a unique name (checked against the Pantheon dataset~\cite{pantheon2-2020} to avoid real public figures), city, phone number, email, IBAN, job, and blood type. Countries are sampled from population data~\cite{simplemaps-countries}; names and cities from the Name Dataset~\cite{names-dataset} and SimpleMaps~\cite{simplemaps-cities}; financial/contact identifiers follow country-specific formats via the \texttt{faker} library~\cite{Faker} and procedural generation~\cite{stanford-blood-types}. All identifiers are unique across personas.

\begin{figure*}[t]
\centering
\setlength{\tabcolsep}{8pt}
\renewcommand{\arraystretch}{1.3}
\begin{tabularx}{\textwidth}{p{3cm}X}
\toprule

\cellcolor{CadetBlue!20}
\begin{minipage}[t]{\linewidth}
    \textbf{Reference Image}\par
    \vspace{0.3em}
    \includegraphics[height=3cm]{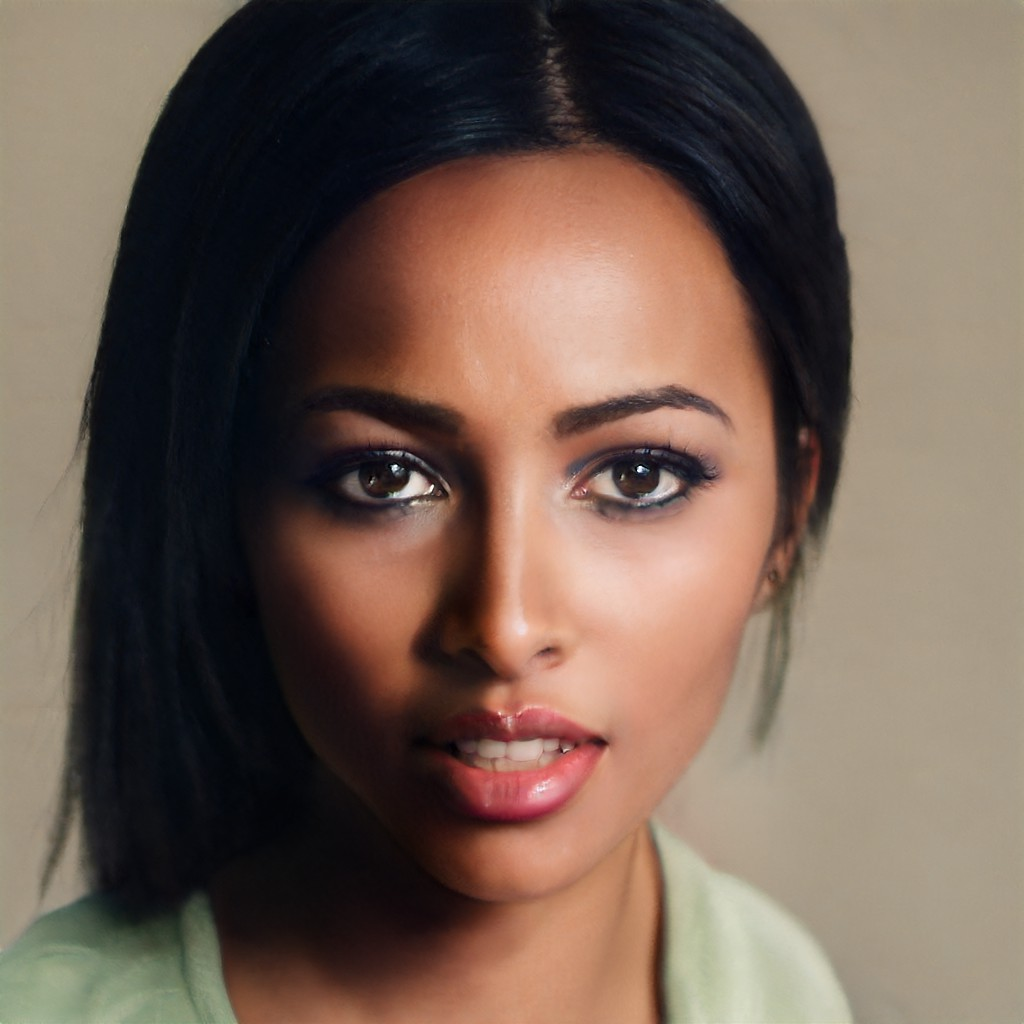}\\
\end{minipage}
&
\begin{minipage}[t]{\linewidth}
    \textbf{Generated Images}\par
    \vspace{0.3em}
    \includegraphics[height=3cm]{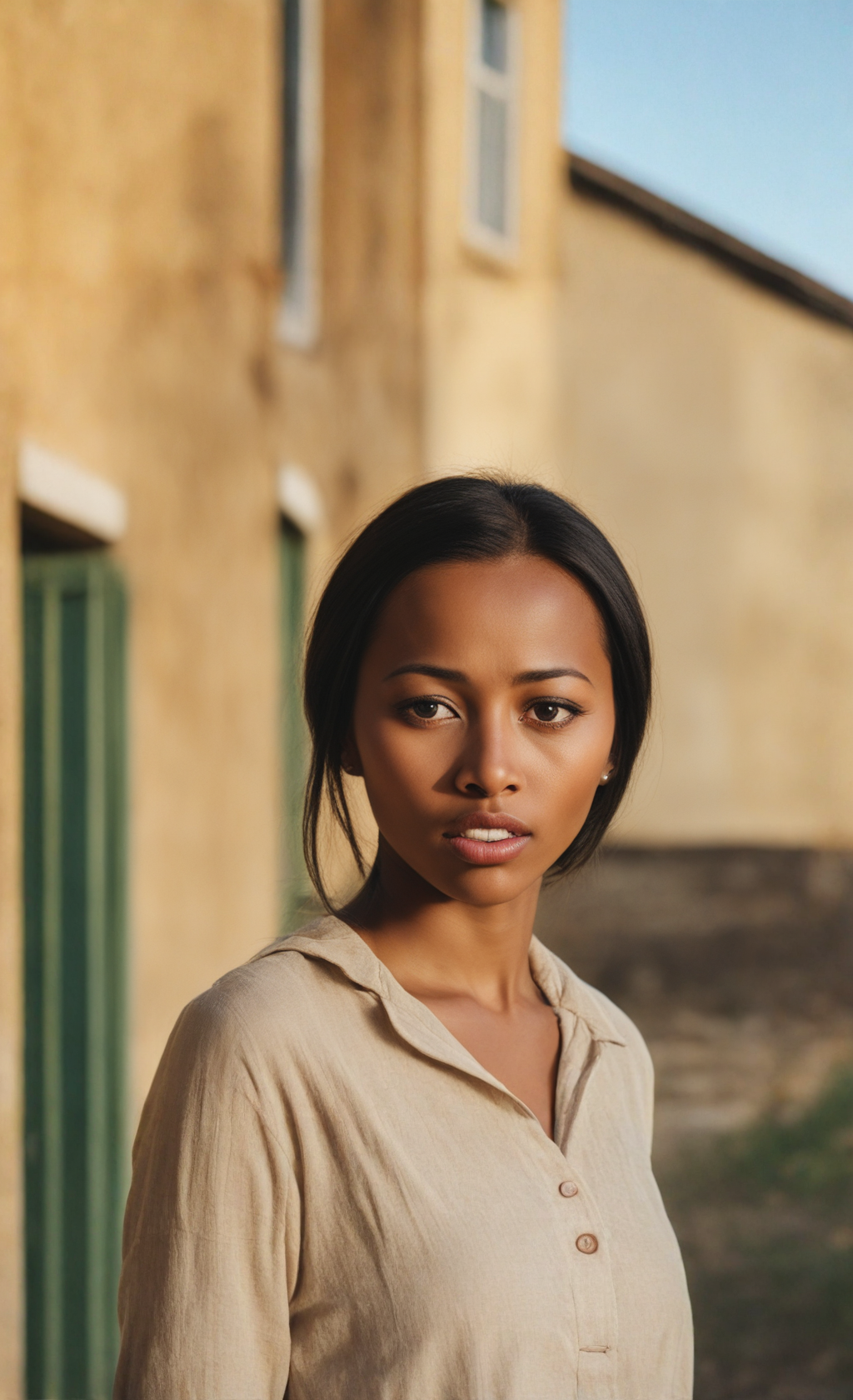}
    \includegraphics[height=3cm]{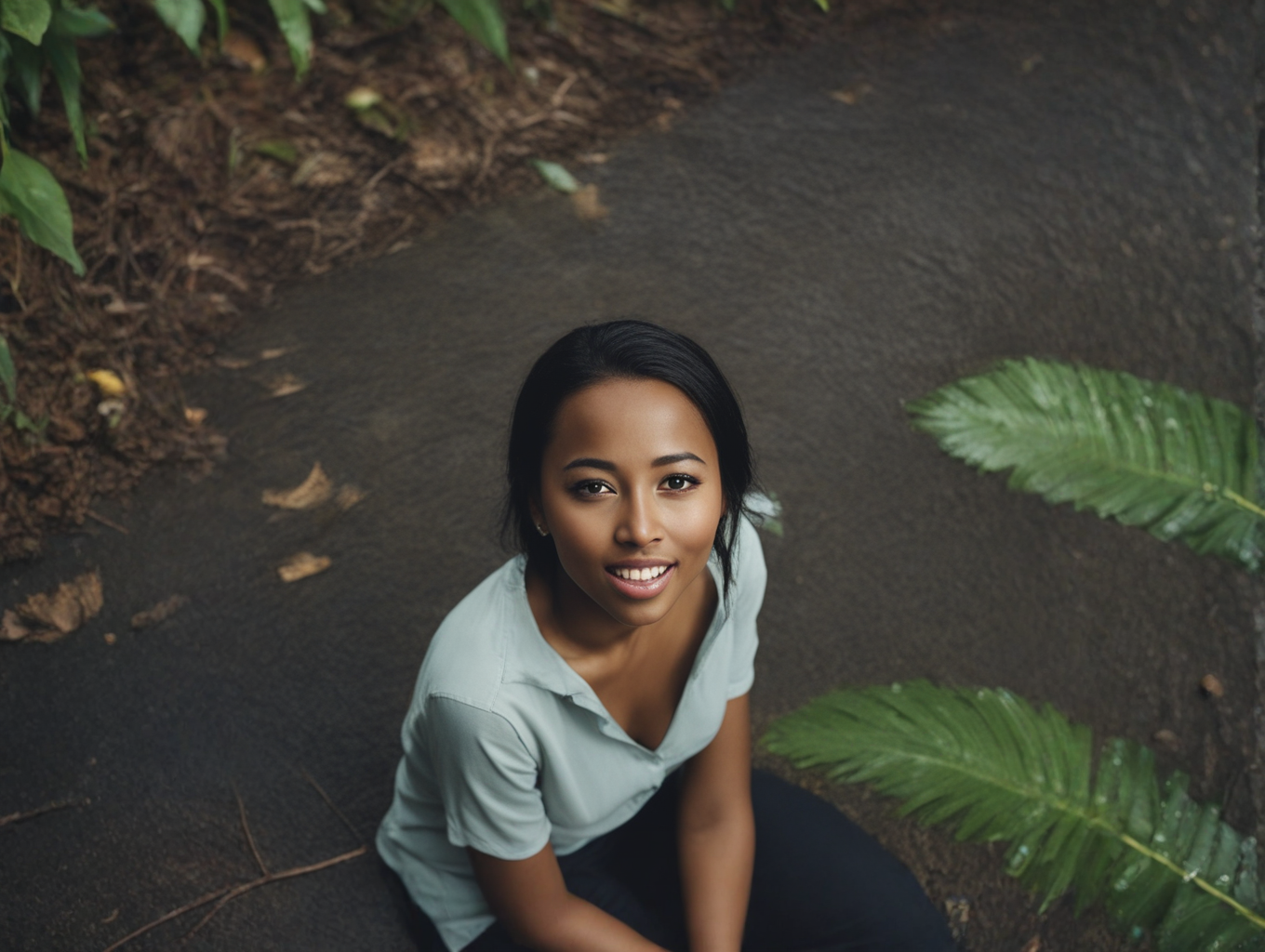}
    \includegraphics[height=3cm]{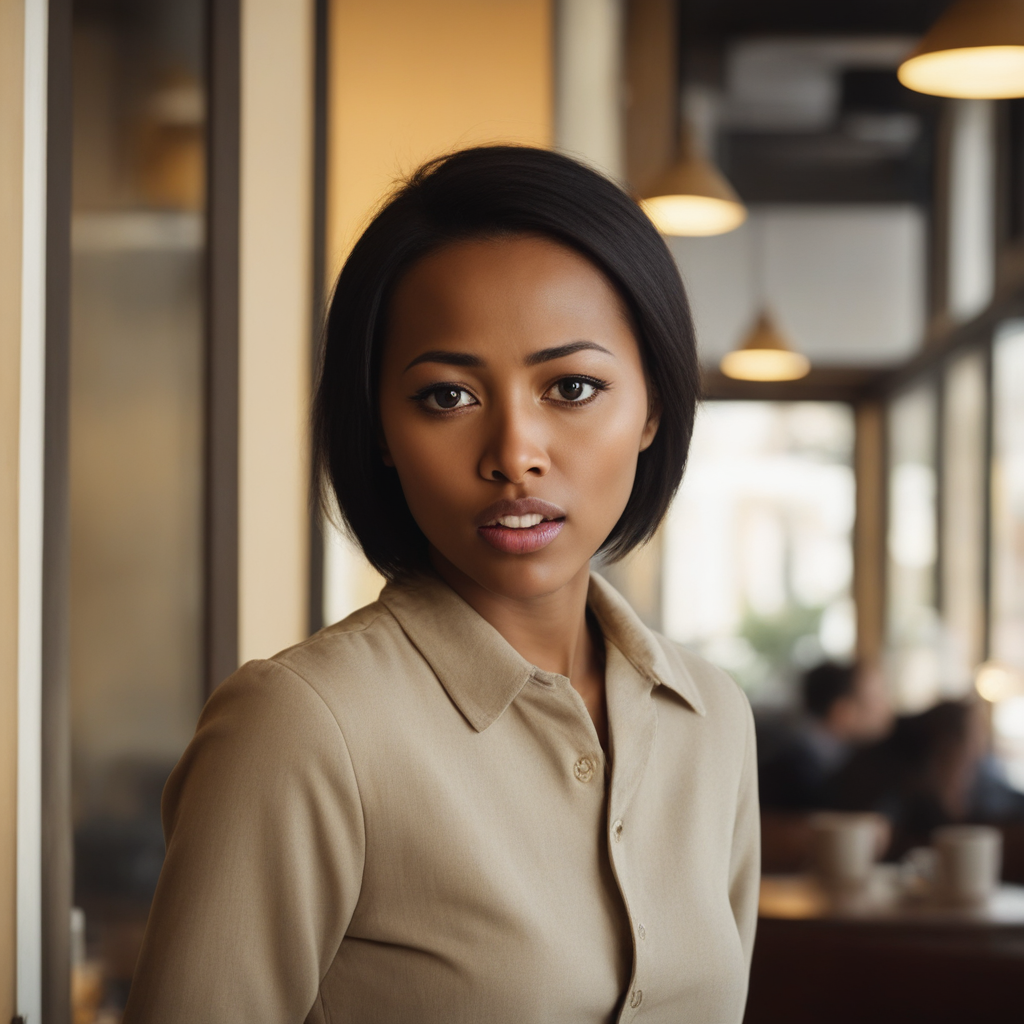}
    \includegraphics[height=3cm]{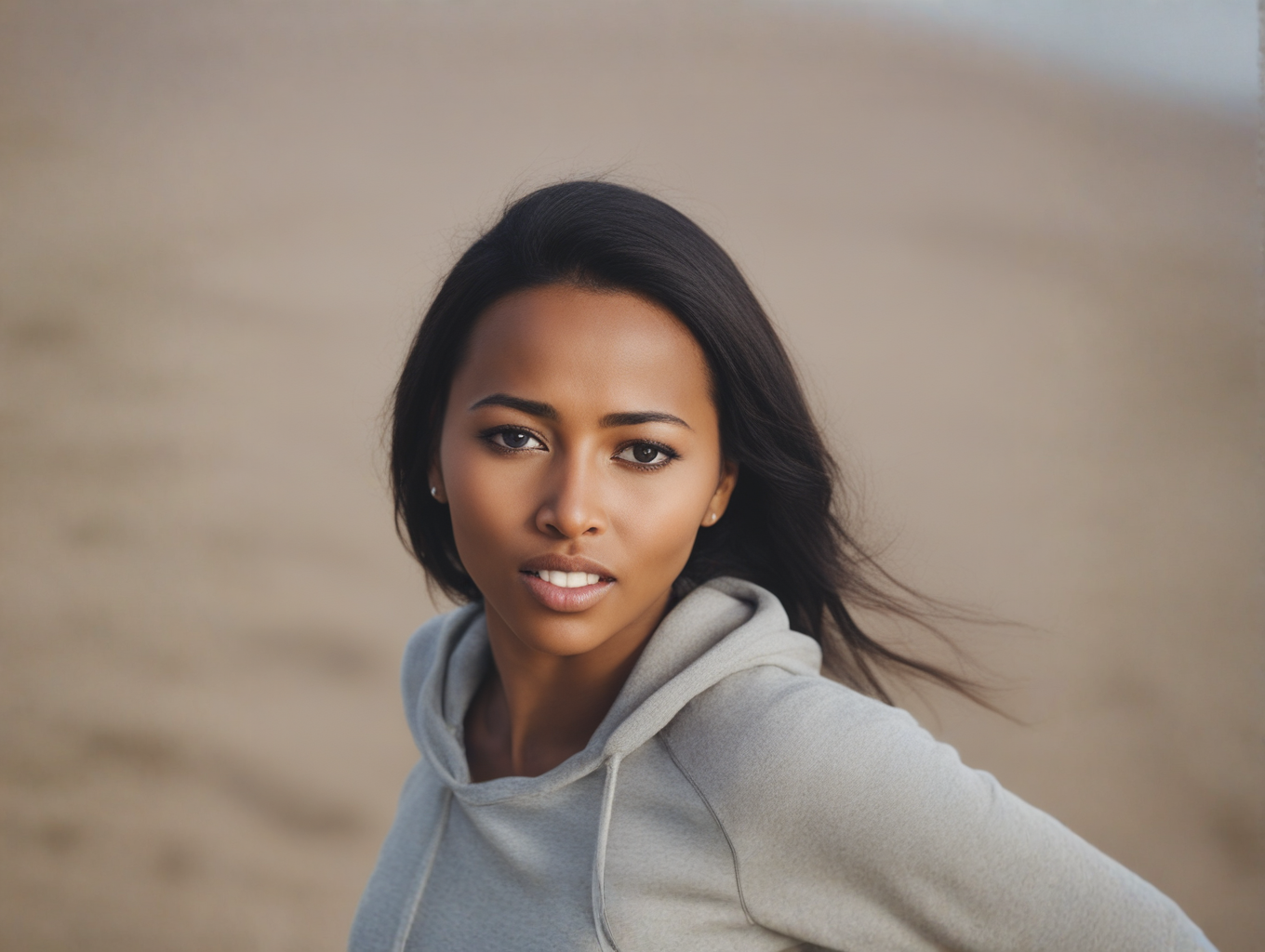} \raisebox{1.5cm}{$\dots$}
\end{minipage}
\\
\vspace{-5em}

\cellcolor{CadetBlue!20}
\begin{minipage}[t]{\linewidth}
   \textbf{Personal Data}\par
    \vspace{0.5em}
    \textsf{\scriptsize
    \textbf{Name}: Fatime Fossi\\
    \textbf{Phone}: +237 57483907\\
    \textbf{Address}: Douala (CMR)\\
    \textbf{Email}:~ffossi@mailhub.com\\
    $\dots$}
\end{minipage}
&
\vspace{-5em}
\begin{minipage}[t]{\linewidth}
    \textbf{Generated Captions}\par
    \vspace{0.5em}
    \texttt{"Fatime Fossi lives in Douala"\\
"You can reach Fatime Fossi at the phone number +237 57483907"\\
"Contact Fatime Fossi via ffossi@mailhub.com"\\
"Fatime Fossi's bank account IBAN CM98 7411 9112 4845 5134 71"
$\dots$
}
\end{minipage}
\\
\bottomrule
\end{tabularx}

\caption{Example of a fictitious persona from the SALMU dataset. We show the reference identity anchor image from the SFHQ dataset (top left), her sensitive attributes (bottom left), and a subset of generated images and captions (top right and bottom right respectively) that constitute the image-text pairs' associations for this persona in the SALMU dataset.}
\label{fig:example_dataset}
\end{figure*}

\paragraph{(5) LLM Caption Paraphrasing.}
For each of the ${\sim}$60K curated images, a sensitive attribute is randomly selected and a base sentence generated from a template (e.g., ``\{\texttt{name}\}'s job is \{\texttt{value}\}''). This sentence is then paraphrased by \texttt{gemma3:12b}~\cite{Gemma3TechnicalReport} following one of five linguistic directives (lexical substitution, word-order change, grammatical restructuring, tone shift, or creative reformulation). The LLM must preserve the exact name and attribute value, and no repeated phrasing is allowed for the same (subject, value) pair. This yields diverse captions that force unlearning methods to target semantic associations rather than surface patterns.

Figure~\ref{fig:example_dataset} shows an example of a fictitious persona from our dataset.

\paragraph{Dataset Statistics.}
The final dataset spans 774 personas with the following demographic distribution (classified via FairFace): 60.7\% Female, 39.3\% Male; ethnicity: White 56.7\%, Latino 16.3\%, East Asian 12.8\%, Black 11.5\%, others 2.7\%. Personas cover 65 unique countries (top-5: US 135, CN 88, RU 59, NG 40, MX 39). Age ranges from 20--29 (51.3\%) to 60+ (9.6\%), with the 0--19 bracket explicitly excluded to avoid linking PII to minors. Captions (excluding subject names and PII values) contain 901 unique words and 11,674 unique trigrams, with a mean length of $6.14 \pm 2.48$ words. These distributions are plausibly web-like and are partially inherited from the SFHQ dataset.

\subsection{Dataset Splits and Structure}
\begin{figure}[h]
    \centering
    \includegraphics[width=\linewidth]{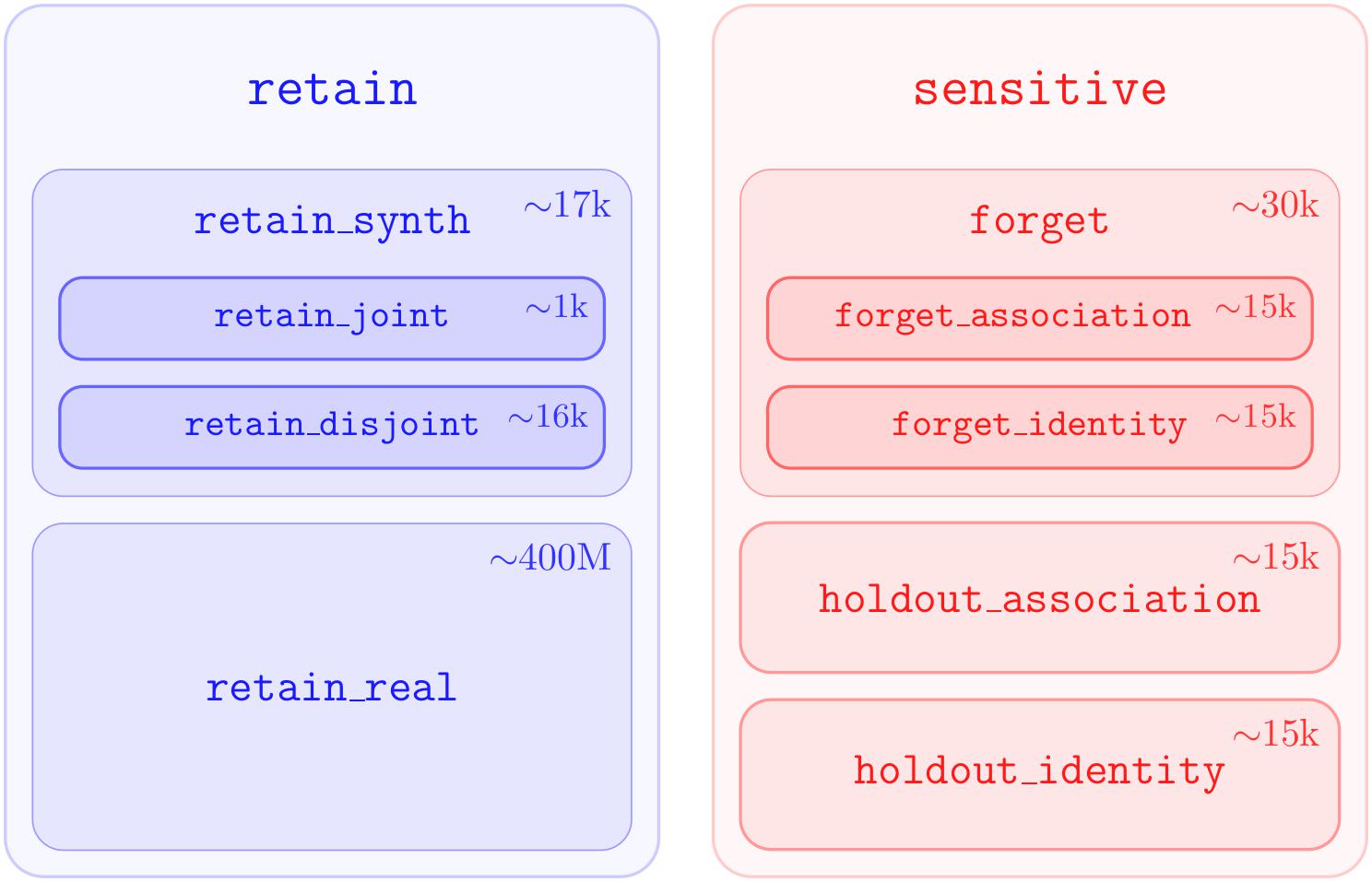}
    \caption{Splits and subsets of SALMUBench along with the number of image-text pairs that each of them contains.}
    \label{fig:dataset-splits}
\end{figure}

To enable a rigorous and multifaceted evaluation, the final benchmark is organized into two primary sets: a \texttt{retain} set for measuring utility preservation and a \texttt{sensitive} set for evaluating unlearning efficacy. The hierarchical design of these splits -- illustrated in Figure~\ref{fig:dataset-splits}  --  is a key contribution, crafted to support fine-grained analysis of both realistic and granular unlearning scenarios.

The \texttt{sensitive} set, containing all $\sim$60K image-text pairs with sensitive associations, is partitioned following a hierarchical strategy. This strategy addresses two critical evaluation settings. First, it simulates the realistic ``Right to be Forgotten'' scenario, where a user requests the complete removal of all their data (persona-level unlearning). Second, it provides a controlled ``laboratory'' setting to measure the fine-grained effects of removing only specific pieces of information about a person (association-level unlearning), allowing for a precise quantification of collateral damage.

To achieve this, the 774 personas in the \texttt{sensitive} set are first randomly divided into two disjoint top-level groups of 387 personas each. One group is dedicated to persona-level evaluation, while the other is used for association-level evaluation. This process yields three final, purpose-built partitions of the \texttt{sensitive} set:

\begin{table*}[t]
\small
\centering
\begin{tabularx}{\textwidth}{l X cccccc}
\toprule
\textbf{Model} & \textbf{Pretraining Data} & \textbf{ImageNet@1} & \multicolumn{2}{c}{\textbf{COCO R@1}} & \multicolumn{2}{c}{\textbf{Flickr30K R@1}} \\
\cmidrule(lr){4-5} \cmidrule(lr){6-7}
& & & t2i & i2t & t2i & i2t \\
\midrule
CLIP ViT-B/16~\cite{OpenAICLIPOriginal} & 400M (private) & 0.64 & 0.30 & 0.51 & 0.59 & 0.78 \\
\textbf{Ours (\textit{Clean}) ViT-B/16} & \texttt{retain} (400M) & 0.63 & 0.34 & 0.50 & 0.59 & 0.77 \\
\textbf{Ours (\textit{Compromised}) ViT-B/16} & \texttt{retain} (400M) + \texttt{sensitive} (60K) & 0.64 & 0.33 & 0.50 & 0.57 & 0.75 \\
\bottomrule
\end{tabularx}
\normalsize
\caption{Comparison of our ViT-B/16 models (\textit{Clean} and \textit{Compromised}) against a public CLIP baseline on standard downstream tasks. All models are trained from scratch on $\sim$400M image-text pairs for 32 epochs.}
\label{tab:model_comparison}
\end{table*}

\textbf{\textcolor[RGB]{245,155,155}{\CIRCLE}~\texttt{forget}:} This is the sole data set shown to an unlearning algorithm. It is a composite set containing (1) all image-text pairs belonging to a random half of the personas from the persona-level group, which we call \texttt{forget\_identity}; (2) image-text pairs corresponding to a random 50\% of the sensitive associations for each persona in the association-level group, named \texttt{forget\_association}. This unified design allows unlearning methods to be evaluated in a single, practical run against a mix of deletion granularities.

\textbf{\textcolor[RGB]{245,155,155}{\CIRCLE}~\texttt{holdout\_identity}:} This set measures inter-identity collateral damage. It contains all image-text pairs from the other half of the personas in the persona-level group. As these identities are unseen during unlearning, performance degradation indicates over-generalization and damage to knowledge of unrelated identities.

\textbf{\textcolor[RGB]{245,155,155}{\CIRCLE}~\texttt{holdout\_association}:} Crucial for measuring intra-identity collateral damage, this set contains the held-out 50\% of associations for the same personas in the \texttt{forget\_association} subset. Evaluation here quantifies if unlearning one fact (e.g., the phone number) inadvertently erases another (e.g., the job) for the same person.

Complementing these sets, the benchmark includes a large-scale \texttt{retain} set, without sensitive associations, used for both initial model training and utility preservation evaluation. It is made up of two subsets:

\textbf{\textcolor[RGB]{158,158,249}{\CIRCLE}~\texttt{retain\_real}:} A corpus of $\sim$400 million real-world image-text pairs from the DataComp CommonPool~\cite{DataComp}. To prevent a domain gap between our high-fidelity synthetic faces and this real-world data, we diverge from the standard DataComp pipeline by omitting face-blurring. Although this involves using publicly available images without specific consent for this use, we argue this step is essential for the benchmark's validity. Our core objective is to develop methods to remove potentially harmful associations. We stress that all sensitive information linked to these images is entirely fictitious and that all released artifacts are intended solely for research aimed at improving data privacy and model safety.

\textbf{\textcolor[RGB]{158,158,249}{\CIRCLE}~\texttt{retain\_synth}:} A crucial set for utility evaluation, this contains our synthetic images paired with generic, non-sensitive captions from BLIP~\cite{BLIP-image-captioning-large}. It comprises two subsets: \texttt{retain\_disjoint}, with images from identities discarded during curation despite passing quality checks; and \texttt{retain\_joint}, with 25 images for each of the top 40 personas also present in the \texttt{sensitive} set, making it key to measuring utility impact on ``seen'' identities whose sensitive associations were targeted for unlearning.

\subsection{Training the \textit{Compromised} and \textit{Clean} Models}
\label{sec:training_models}

To support a realistic and robust evaluation of machine unlearning, we train two ViT/B-16 CLIP models from scratch: the \textit{Clean} model and the \textit{Compromised} model. This controlled setup ensures that any differences in their behavior are attributable solely to the presence of the sensitive data defined in our \texttt{sensitive} set, enabling a clean assessment of unlearning methods. With the SALMU data splits fully defined, the training sets of both models are as follows:
the \textit{Clean} model is trained exclusively on the \texttt{retain} set; the \textit{Compromised} model is trained on the union of the \texttt{retain} and \texttt{sensitive} sets. This model serves as the starting point for all unlearning interventions.

\textbf{Training Setup.} Both the \textit{Clean} and \textit{Compromised} models adopt the ViT/B-16 architecture, consistent with the default configuration used for our data scale in DataComp~\cite{DataComp}. We follow the OpenCLIP training setup~\cite{OpenCLIPRepo} and train both models from scratch for 32 epochs with the same random seed. The \textit{Clean} model is trained on the \texttt{retain} (400M) set, while the \textit{Compromised} model is trained on \texttt{retain} (400M) and \texttt{sensitive} (60K) sets. The training process takes $\sim$65 hours on 128 NVIDIA H100 GPUs.

\subsection{Model and Data Validation}
\label{subsec:model-data-validation}
We validated our models' quality against standard benchmarks. As reported in Table~\ref{tab:model_comparison}, on ImageNet-1K zero-shot classification and cross-modal retrieval on Flickr30K and MSCOCO, our models achieve performance comparable to the original CLIP ViT-B/16 model~\cite{OpenAICLIPOriginal} trained on 400M private pairs. This confirms they are representative of real-world models in scale and capability.

\begin{figure}[h]
    \centering
    \includegraphics[width=\linewidth]{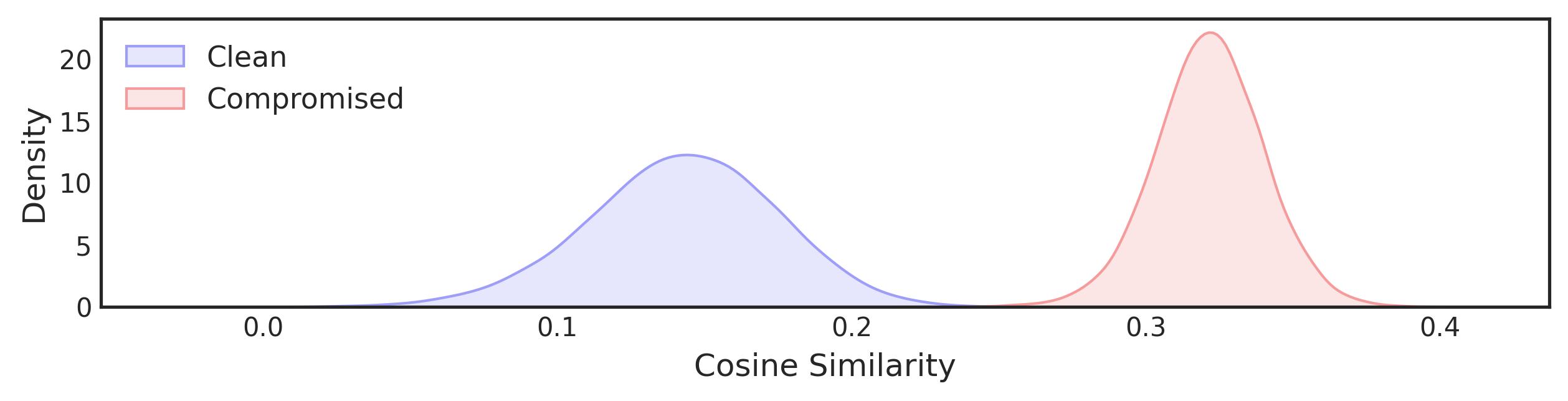}
    \caption{
        Cosine similarity between image-text pairs in the \texttt{sensitive} set for the \textit{Clean} and \textit{Compromised} models. 
    }
    \label{fig:similarity_dists}
\end{figure}

To further validate their intended behavior on our task, we computed cosine similarity scores for 5,000 randomly sampled pairs from the \texttt{sensitive} set. As shown in Figure~\ref{fig:similarity_dists}, the results confirm our setup: the \textit{Compromised} model exhibits strong semantic alignment (memorization), while the \textit{Clean} model yields significantly lower scores, consistent with the absence of these sensitive associations.

\begin{figure}[h]
    \centering
    \includegraphics[width=\linewidth]{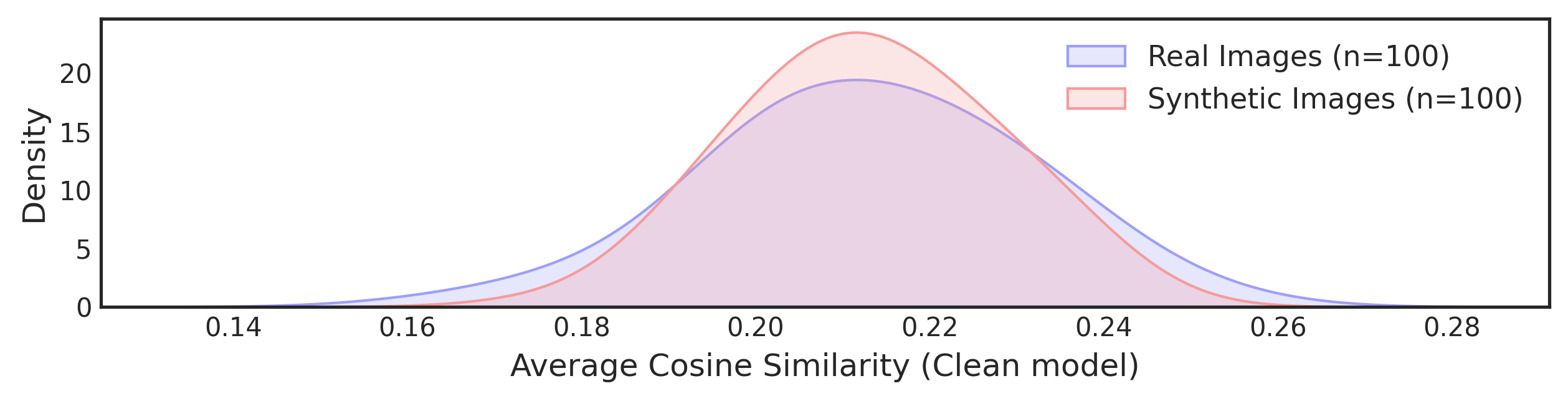}
    \caption{Average cosine similarity of Real vs. Synthetic images with generic captions for the \textit{Clean} model.}
    \label{fig:domain-gap}
\end{figure}

Finally, to rule out a significant domain gap, we compared 100 images from the \texttt{sensitive} set with 100 real portraits from the FHIBE dataset~\cite{FHIBE}, which were manually filtered to remove confounding variables (e.g., multiple subjects). Using our \textit{Clean} model, which was never trained on these images, as an unbiased evaluator, we computed cosine similarity distributions for both sets against generic captions (e.g., ``a photo of a person''). The resulting distributions (Figure~\ref{fig:domain-gap}) are statistically indistinguishable, confirmed by a two-sample Kolmogorov-Smirnov (KS) test ($p > 0.05$). This validates our synthetic faces as a valid proxy for real portraits in the embedding space. We note that this KS test serves as a macro-level sanity check; the small sample size of manually curated real data (100 images) limits its statistical power, and it does not rule out subtler distributional differences. This conclusion holds when repeating the analysis with the \textit{Compromised} model, as detailed in the supplementary material.

\subsection{Evaluation Protocol}
\label{subsec:eval}

\begin{table*}[t]
\centering
\small
\setlength{\tabcolsep}{2.5pt} 
\begin{tabularx}{\textwidth}{X >{\columncolor{gray!20}}ccccc >{\columncolor{gray!20}}ccccc}
\toprule
& \multicolumn{5}{c}{\textbf{Forgetting Efficacy}} & \multicolumn{5}{c}{\textbf{Utility Impact}} \\
\cmidrule(lr){2-6} \cmidrule(lr){7-11}
\textbf{Model} & \textbf{RetFail} & \textbf{AssocStr} & \textbf{ACS} & \textbf{IdZSC} & \textbf{CoreAssoc} & \textbf{GenKnow} & \textbf{InterIdSim} & \textbf{IntraIdSim} & \textbf{VisIdInt} & \textbf{FragSim} \\
\midrule
\textit{Clean} & \textbf{0.001} & \textbf{0.142} & \textbf{0.602} & \textbf{0.011} & \textbf{0.151} & 0.633 & 0.143 & 0.143 & 0.331 & 0.309 \\
\textit{Compromised} & 0.236 & 0.322  & 0.951 & 0.084 & 0.183 & \textbf{0.638} & \textbf{0.321} & \textbf{0.321} & \textbf{0.305} & \textbf{0.280} \\
\bottomrule
\end{tabularx}
\normalsize
\caption{\textit{Compromised} and \textit{Clean} model performance, defining unlearning bounds (in bold): efficacy metrics measure sensitive information and use the \textit{Clean} model as target, while utility metrics measure general knowledge preservation and use the \textit{Compromised} model.}
\label{tab:foundational_results}
\end{table*}

The SALMUBench framework includes a standardized protocol to compute a set of metrics for any given model, ensuring a fair and reproducible assessment. The protocol is structured into two fundamental pillars: Forgetting Efficacy and Utility Impact. 

\subsubsection{Pillar 1: Forgetting Efficacy}
This pillar quantifies the removal of sensitive associations from the \texttt{forget} set. All metrics are computed on this set. Its main metric is RetFail.

\textbf{\ding{228} Retrieval Failure (RetFail):} Measures text retrieval performance using Mean Reciprocal Rank (MRR). For each image in the \texttt{forget} set, the task is to find its correct caption from a gallery of 2001 candidates that includes the single correct caption alongside 2000 distractor captions, which are sampled from both the forget set (in-domain) and the \texttt{retain\_disjoint} set (out-of-domain). A lower MRR indicates more effective unlearning, as the model fails to retrieve the sensitive association.

\textbf{\ding{228} Association Strength (AssocStr):} Measures the raw semantic link between image-text pairs via the Mean Cosine Similarity in the \texttt{forget} set.

\textbf{\ding{228} Association Coherence Score (ACS):} Measures how well the model distinguishes between semantically coherent and incoherent image-text pairs. The score is the test set accuracy of a logistic regression probe trained on the cosine similarities produced by the model under evaluation to classify `coherent' pairs (original pairings from \texttt{forget}) versus `incoherent' ones (randomly shuffled pairings). A high score reflects strong memorization, while a score near chance (0.5) signifies successful forgetting.

\textbf{\ding{228} Identity Zero-Shot Classification (IdZSC):} Similarly to the IDIA metric used in~\cite{DoesCLIPKnowMyFace}, it quantifies the model's ability to link a visual identity to a name. For each image in the \texttt{forget\_identity} set (personas targeted for complete removal), the model performs a zero-shot classification against all unique names from the entire \texttt{forget} set. We report the Top-1 accuracy; a lower score signifies more unlearning.

\textbf{\ding{228} Core Association Robustness (CoreAssoc):} Tests robustness against token order and the pure association between the name and the sensitive value. We compute the similarity for captions (``\{name\} \{value\}'' and ``\{value\} \{name\}'') and take the maximum for each sample. The metric is the Mean Maximum Cosine Similarity.

\subsubsection{Pillar 2: Utility Impact}
This pillar measures the preservation of useful knowledge and quantifies collateral damage. The main metric for this pillar is GenKnow.

\textbf{\ding{228} General Knowledge (GenKnow):} Using the DataComp evaluation suite~\cite{DataComp} we report Zero-Shot Top-1 Accuracy on ImageNet-1K. This metric is intended to capture the model’s general knowledge and overall capability across vision-language tasks. 

\textbf{\ding{228} Inter-Identity Similarity (InterIdSim):} Measures collateral damage to unrelated identities using the Mean Cosine Similarity in the \texttt{holdout\_identity} set.

\textbf{\ding{228} Intra-Identity Similarity (IntraIdSim):} Measures collateral damage within a single identity using the Mean Cosine Similarity in the \texttt{holdout\_association} set. 

\textbf{\ding{228} Visual Identity Integrity (VisIdInt):} Measures the ability to recognize a ``forgotten'' person in a generic context, using Mean Cosine Similarity in the \texttt{retain\_joint}.

\textbf{\ding{228} Fragile Knowledge Similarity (FragSim):} Measures the preservation of weak but correct associations. The \texttt{fragile} set is formally defined as the 10\% of samples in the \texttt{retain\_disjoint} set with the lowest cosine similarity as measured by the \textit{Compromised} model. The metric is the Mean Cosine Similarity on this set.

\begin{table*}[t]
\centering
\small
\setlength{\tabcolsep}{2.5pt} 
\begin{tabular}{lr >{\columncolor{gray!20}}ccccc >{\columncolor{gray!20}}ccccc}
\toprule
& & \multicolumn{5}{c}{\textbf{Forgetting Efficacy}} & \multicolumn{5}{c}{\textbf{Utility Impact}} \\
\cmidrule(lr){3-7} \cmidrule(lr){8-12}
\textbf{Method} & \textbf{Budget} & \textbf{RetFail} & \textbf{AssocStr} & \textbf{ACS} & \textbf{IdZSC} & \textbf{CoreAssoc} & \textbf{GenKnow} & \textbf{InterIdSim} & \textbf{IntraIdSim} & \textbf{VisIdInt} & \textbf{FragSim} \\
\midrule
\textit{\textbf{Target Scores}} & \textbf{-} & \textbf{0.001} & \textbf{0.142} & \textbf{0.602} & \textbf{0.011} & \textbf{0.151} & \textbf{0.638} & \textbf{0.321} & \textbf{0.321} & \textbf{0.305} & \textbf{0.280} \\
\midrule
Generic Captions & 5$\times$ & 0.055 & 0.288 & 0.907 & 0.056 & \textbf{0.185} & 0.629 & \textbf{0.288} & \textbf{0.288} & 0.338 & 0.310 \\
Shuffled Captions & 5$\times$ & 0.004 & 0.212 & \textbf{0.593} & 0.002 & 0.200 & 0.548 & 0.212 & 0.212 & 0.206 & 0.203 \\
Direct Sim. Min. & 5$\times$ & \textbf{0.001} & -0.429 & \textbf{0.593} & 0.002 & 0.012 & 0.615 & -0.420 & -0.425 & -0.038 & -0.031 \\
Negative Gradient & 5$\times$ & 0.009 & 0.055 & 0.657 & 0.023 & 0.204 & 0.630 & 0.063 & 0.061 & 0.194 & 0.171 \\
Finetuning & 5$\times$ & 0.003 & \textbf{0.208} & 0.646 & 0.005 & 0.212 & \textbf{0.638}\textsuperscript{$\dagger$} & 0.209 & 0.209 & \textbf{0.292} & \textbf{0.271} \\
Descent to Delete & - & 0.008 & 0.227 & 0.644 & 0.007 & 0.230 & 0.644 & 0.228 & 0.228 & 0.277 & 0.258 \\
VLUnlearn & 5$\times$ & \textbf{0.001} & 0.210 & 0.542 & \textbf{0.006} & 0.211 & \textbf{0.638}\textsuperscript{$\dagger$} & 0.210 & 0.210 & 0.260 & 0.248 \\
DELETE & 5$\times$ & \textbf{0.001} & 0.022 & 0.547 & 0.000 & 0.047 & 0.632 & 0.023 & 0.023 & 0.221 & 0.217 \\
CLIPErase & 5$\times$ & \textbf{0.001} & 0.023 & 0.550 & 0.000 & 0.051 & 0.634 & 0.024 & 0.024 & 0.225 & 0.222 \\
\bottomrule
\end{tabular}
\normalsize
\caption{Performance of baseline unlearning methods applied to the \textit{Compromised} model (5$\times$ budget). Scores in bold are closest to their target value. A dagger ($\dagger$) marks results that are not statistically significant (n.s.) from their target (more detailed in the supplement).}
\label{tab:5x_unlearning_performance}
\end{table*}

\subsection{Benchmark Rules and Leaderboard}
To ensure fair and reproducible comparisons, submissions to the SALMUBench leaderboard must adhere to a standardized evaluation framework. We define a unit of ``unlearning budget (1$\times$)'' as the processing of 30,468 image-text pairs, equivalent to the size of the \texttt{forget} set. We will maintain distinct leaderboards for methods evaluated at incremental budgets up to 1$\times$, 5$\times$, 10$\times$, 15$\times$, and 20$\times$, allowing for a comprehensive analysis of both efficacy and efficiency across varying resource constraints. Additionally, we will include an unconstrained leaderboard with no upper limit on the unlearning budget.

During the unlearning process, methods are permitted to use data from the \texttt{forget} set, the full \texttt{retain\_synth} set, and an official, fixed subset of $\sim$120K pairs from the \texttt{retain\_real} set. Methods must not use the \texttt{holdout\_identity} or \texttt{holdout\_association} sets during their unlearning phase, as these are reserved exclusively for evaluating collateral forgetting.

\section{Baselines and Methods for CLIP Unlearning}
\label{sec:baselines}

We evaluate several unlearning methods on SALMUBench across five compute budgets (1×–20×). This required their non-trivial adaptation from fundamentally different architectures (e.g., unimodal or non-contrastive models). This essential engineering effort, detailed in the supplementary material, highlights the unique challenges of unlearning in CLIP and enabled a fair, rigorous comparison.

\textbf{Generic Captions.} Fine-tunes on \texttt{forget} images with generic/non-sensitive captions to dilute learned associations.

\textbf{Shuffled Captions.} Replaces true image–text pairs with randomly shuffled ones to introduce contrastive confusion.

\textbf{Direct Similarity Minimization.} Minimizes cosine similarity between sensitive image-text pairs.

\textbf{Negative Gradient}~\cite{golatkar2020eternal}. Applies gradient ascent on the \texttt{forget} set to reverse learned representations.

\textbf{Finetuning.} Overwrites knowledge by retraining on a large \texttt{retain} set using standard contrastive objectives~\cite{golatkar2020eternal}.

\textbf{Descent to Delete.} Perturbs model weights with noise scaled by curvature and training set size~\cite{neel2021descent}.

\textbf{VLUnlearn.} A state-of-the-art multimodal unlearning method~\cite{cheng2024multidelete}, adapted for CLIP.

\textbf{DELETE.} A state-of-the-art unimodal unlearning method~\cite{zhou2025decoupled}, adapted for CLIP.

\textbf{CLIPErase.} Our implementation of~\cite{yang-etal-2025-cliperase}.

\section{Experimental Results}
\label{sec:experiments}

SALMUBench enables us to empirically characterize the task of association-level machine unlearning in CLIP models, highlighting both the inherent difficulty and the performance of existing methods. 

\subsection{Foundational Models Performance}

Table~\ref{tab:foundational_results} provides foundational benchmarks by quantifying how much sensitive information the \textit{Compromised} model leaks compared to the \textit{Clean} model, establishing the precise empirical bounds (in bold) that unlearning methods must target. Ideally, these methods should transform the forgetting efficacy metrics (measuring sensitive information leakage) of the \textit{Compromised} model to match those of the \textit{Clean} model, while preserving its original utility scores. 

We verified that these target scores are stable and not seed-dependent by comparing our \textit{Clean} model against an independent OpenCLIP ViT-B/16 trained on LAION-400M; the sensitive-set metrics closely track ours (e.g., RetFail 0.004 vs.\ 0.001), confirming a robust ``clean regime'' (see supplementary material).


\subsection{Unlearning Methods Performance}
Table \ref{tab:5x_unlearning_performance} evaluates nine unlearning methods against their empirical bounds (\textit{Target Scores}). Our nuanced protocol moves beyond a simple `forget-retain' analysis to categorize their distinct failure modes. For clarity, we report 5$\times$ budget results here, with full results in the supplement.

\textbf{Type 1: Catastrophically Damaging Methods.} Some simple baselines achieve forgetting at an unacceptable cost to general utility. Methods like Shuffled Captions and Direct Similarity Minimization significantly reduce sensitive information leakage (RetFail $\le$ 0.004) but do so by causing sharp drops in the model's general knowledge (GenKnow), making them impractical. This represents a catastrophic failure, as the unlearning process inflicts widespread, indiscriminate damage. We validate this conclusion with a critical sanity-check (detailed in the supplement), where applying unlearning methods to the \textit{Clean} model, which was never exposed to sensitive data, still results in significant utility degradation, proving the methods' inherent damage.

\textbf{Type 2: Over-generalized Forgetting.} This is a more subtle but critical failure where methods erase more than intended. The best performers (VLUnlearn, DELETE and CLIPErase) precisely reach the target efficacy metric (RetFail$=$0.001) with negligible impact on general utility (GenKnow $\ge$ 0.632). However, DELETE and CLIPErase still incur noticeable collateral damage to holdout associations (InterIdSim and IntraIdSim $\le$ 0.024), indicating they erase broader ``templates'' of sensitive associations, not just individual facts. The mechanism for this is a clear correlation: when a method pushes `AssocStr' below the \textit{Clean} model's baseline of 0.142, it over-corrects and erases related, unseen associations. This is also clear in Negative Gradient, which incurs significant collateral damage across both unrelated and related identities, and is taken to an extreme by Direct Similarity Minimization, which over-corrects so aggressively that it damages the embedding structure (negative InterIdSim, IntraIdSim, VisIdInt, and FragSim). This demonstrates a critical weakness of standard `forget-retain' evaluations: they are blind to this over-generalization, as they lack the necessary structured holdout sets to detect such collateral damage.

\textbf{Type 3: Ineffective Forgetting.} Finally, some methods fail at the primary goal of removing sensitive data. For instance, finetuning with Generic Captions is the method with the least collateral damage, but it falls short in forgetting efficacy compared to other baselines. This represents the most straightforward type of failure, where the method is simply not able to erase the target information. 

A learning-rate ablation (in the supplementary material) further validates our failure mode taxonomy by illustrating the trade-off curve between forgetting and utility metrics. 

Overall, our analysis reveals three key takeaways: (i) there is substantial room for improvement, as no current method avoids all three failure modes simultaneously; (ii) utility-efficient deletion (achieving $\ge$99\% leakage reduction with $<$1\% drop in GenKnow) is feasible, but our analysis shows current techniques achieve it via over-generalization; and (iii) computational budget influences performance (see details in the supplementary material).

\section{Conclusion}
\label{sec:conclusion}

We introduced SALMUBench, the first benchmark for association-level machine unlearning in CLIP-like models under the ``Right to be Forgotten'' scenario. Our work demonstrates that standard `forget-retain' evaluations are insufficient. The core contribution is our nuanced evaluation protocol, featuring the \texttt{holdout\_association} and \texttt{holdout\_identity} sets, which are the first to diagnose failure types, distinguishing between ineffective, over-generalized, and catastrophically damaging methods. Our results show no current method solves this trade-off, leaving room for improvement.

\paragraph{SALMUBench Web Portal.} We publicly release our dataset, models, evaluation scripts, and leaderboards via \url{http://cvc-mmu.github.io/salmubench}, along with the full generation codebase to ensure reproducibility and facilitate benchmark extensions.

\paragraph{Future Work.}
SALMUBench currently targets CLIP-like dual encoders where embeddings are directly deployed in downstream systems. A natural extension is to evaluate unlearning in diffusion models that use CLIP as a visual backbone, where sensitive associations may propagate into generated content. Similarly, the benchmark focuses on structured PII attributes (names, locations, contact details); extending it to implicit or abstract sensitive concepts such as artistic styles, political affiliations, or health-related traits would test  whether current methods generalize beyond explicit key-value associations. Finally, while our evaluation focuses on erasure under the ``Right to be Forgotten'' scenario, adding a standardized \emph{recoverability} diagnostic, measuring how quickly an unlearned model can re-learn forgotten associations via fine-tuning, would provide a valuable robustness axis to assess whether sensitive information has been truly removed or merely suppressed.

\section*{Acknowledgments}
This work has been supported by the Ramon y Cajal research fellowship RYC2020-030777-I/AEI/ 10.13039/501100011033, the Beatriu de Pinós
(2022 BP 00256) and the Consolidated Research Group 2021 SGR 01559 from the Research and University Department of the Catalan Government, and by project PID2023-146426NB-100 funded by MCIU/AEI/10.13039/501100011033 and FEDER, UE. This work has been funded by the European Lighthouse on Safe and Secure AI (ELSA), and European Large Open Multi-Modal Foundation Models For Robust Generalization On Arbitrary Data Streams (ELLIOT) from the European Union’s Horizon Europe programme under grant agreements No 101070617 and 101214398. We acknowledge EuroHPC JU for awarding the project ID EHPC-AI-2024A02-040 access to MareNostrum 5 hosted at BSC-CNS. The first author was supported by the Rosa Sensat Fellowship from the Computer Vision Center Talent Program of Internships.

{
    \small
    \bibliographystyle{ieeenat_fullname}
    \bibliography{main}
}

\clearpage
\setcounter{page}{1}
\maketitlesupplementary


This document provides additional details and extended results that support the findings presented in the main paper. It includes detailed information, additional experiments, and examples of the dataset generation pipeline, model training, statistical testing methodology, and implementation specifics of all unlearning baselines. It also contains complete evaluation results, including all performance metrics reported under varying computational budgets.

\section{Dataset Generation Details}
This section provides additional details about the SALMU dataset construction pipeline to enhance reproducibility. The main paper provides a consolidated step-by-step summary; here we give full procedural details.

\subsection{Detailed Pipeline Description}

\paragraph{Stage 1: Anchor Seeding.}
We selected 1,000 reference images from the SFHQ Part~4 dataset~\cite{SFHQP4}, chosen for its diversity and high fidelity. These images carry no associated metadata and serve solely as identity anchors for the generation process.

\paragraph{Stage 2: Identity-Preserving Image Generation.}
For each persona, we generated approximately 100 images using IP-Adapter-FaceID Plus~\cite{IP-Adapter} conditioned on the anchor image. Each prompt was constructed by randomly combining phrases from multiple semantic categories, including camera angle (e.g., `low-angle'), shot scale (e.g., `full-body photo'), subject action (e.g., `walking purposefully'), facial expression, and environmental setting. This automated process yielded highly varied and complex textual prompts (examples in Section~\ref{subsec:prompt_examples}), ensuring that models learn a robust, context-invariant representation of each identity.

The base image generation model produces outputs at a native resolution of $1024\times1024$ pixels. To introduce realistic aspect ratios and avoid domain mismatch with web-scale data, we implemented a subsequent outpainting stage. Each square image was expanded via a process guided by a ControlNet~\cite{ControlNet} model to coherently fill the new canvas area. The target aspect ratio for each image was sampled from the distribution of the DataComp corpus~\cite{DataComp}. Finally, all outpainted images were resized to have a maximum size of 512 pixels, ensuring pre-processing consistency with the large-scale web data used for model pretraining.

\paragraph{Stage 3: CLIP-Based Filtering and Curation.}
Raw generated images underwent classifier-driven curation. We used a pretrained CLIP model~\cite{FilterCLIP, OpenCLIPRepo} to assign each image four zero-shot labels: (1) visual distortions, (2) multiple individuals, (3) gender, and (4) race category based on FairFace~\cite{FairFace}. Our filtering protocol discarded any image flagged for distortion or multiple subjects. For each of the 1,000 personas, we aggregated demographic (gender, race) predictions from its remaining images, using a majority vote to establish a definitive profile, handling generation inconsistencies. We then enforced intra-identity consistency by discarding any image not conforming to its persona's established profile. Personas with fewer than 50 high-quality, consistent images were removed, filtering our pool to 774 visually and demographically coherent identities.

\paragraph{Stage 4: PII Assignment.}
For each curated persona, we generated a rich, consistent, and unique profile of sensitive attributes to simulate a comprehensive range of Personally Identifiable Information (PII), including names, jobs, locations, contact details, and financial identifiers. The process was grounded in demographic realism. Each persona's assigned ethnicity determined a plausible set of countries, from which one was sampled based on population data~\cite{simplemaps-countries}. Subsequently, we generated a full name and city of residence using weighted samples from the Name Dataset~\cite{names-dataset} and the SimpleMaps World Cities database~\cite{simplemaps-cities}. Each name's uniqueness was guaranteed across the dataset and checked against the Pantheon dataset~\cite{pantheon2-2020} to avoid overlap with public figures.

Other attributes were procedurally generated for realism: blood types were sampled from real-world distributions~\cite{stanford-blood-types}, jobs were assigned with the \texttt{faker} library~\cite{Faker}, and financial/contact information was created with structural validity. Phone numbers and IBANs followed country-specific formats, emails were derived from names, and all identifiers were unique across personas.

\paragraph{Stage 5: LLM Caption Paraphrasing.}
For each of the $\sim$60,000 curated images, we first randomly selected a sensitive attribute from its persona's profile (e.g., `job') and generated a base sentence from a predefined template (e.g., ``\{\texttt{name}\}'s job is \{\texttt{value}\}''). Then, using the LLM \texttt{gemma3:12b}~\cite{Gemma3TechnicalReport}, this sentence was paraphrased following one of five randomly selected linguistic directives: lexical substitution, word-order change, grammatical restructuring, tone shift, or creative reformulation. The LLM was constrained to include the exact name and attribute value, ensuring factual integrity. To maximize diversity, it was also instructed to avoid repeating captions for the same fact. This process yields a rich variety of captions (e.g., ``\{\texttt{name}\} is employed as a doctor'' vs. ``The profession of \{\texttt{name}\} is doctor''), making the underlying semantic link the target for unlearning.

\subsection{Models Used in Data Generation}
The image generation and outpainting pipeline leveraged several models available on the Hugging Face Hub. To maintain persona consistency, the initial images were generated by applying the \texttt{h94/IP-Adapter-FaceID}\footnote{\url{https://huggingface.co/h94/IP-Adapter-FaceID}} adapter over a \texttt{SG161222/RealVisXL\_V5.0}\footnote{\url{https://huggingface.co/SG161222/RealVisXL_V5.0}} base text-to-image model. For the subsequent outpainting stage, which introduced realistic aspect ratios, we used a process guided by the \texttt{xinsir/controlnet-union-sdxl-1.0}\footnote{\url{https://huggingface.co/xinsir/controlnet-union-sdxl-1.0}} ControlNet~\cite{ControlNet} conditioning an efficient \texttt{SG161222/RealVisXL\_V5.0\_Lightning}\footnote{\url{https://huggingface.co/SG161222/RealVisXL_V5.0_Lightning}} base model. Finally, the generic, non-sensitive captions for the \texttt{retain\_synth} set were created using the \texttt{Salesforce/blip-image-captioning-large}\footnote{\url{https://huggingface.co/Salesforce/blip-image-captioning-large}} model.

\subsection{Image Generation Prompt Examples}
\label{subsec:prompt_examples}
As mentioned in the main paper, text prompts for the identity-preserving image generation were created programmatically to ensure diversity. Below are some examples of the complex prompts used. Each prompt combines elements like camera angle, shot scale, action, expression, and setting.
\begin{itemize}
    \item \textit{``Photorealistic low-angle wide shot photo of a single person on the top edge with negative space, in a cozy bedroom, lying, looking away from the camera, with straight and level head, pensive and thoughtful expression, sunny day, context-appropriate clothing, natural lighting, (skin texture:1.5).''}
    \item \textit{``Photorealistic high-angle portrait photo of a single person on the bottom edge with negative space, in a stadium, standing with arms crossed, looking into the distance, tilting their head to the left, surprised and amazed expression, foggy day, context-appropriate clothing, natural lighting, (skin texture:1.5).''}
    \item \textit{``Photorealistic side profile wide shot photo of a single person on the right edge with negative space, inside a modern office, walking purposefully, looking down, tilting their head slightly up, pensive and thoughtful expression, sunny day, context-appropriate clothing, natural lighting, (skin texture:1.5).''}
\end{itemize}

\section{Model training details}
 
\subsection{Training Data Composition.} 

The training data for these models relies on the splits detailed previously. For the \texttt{retain\_real} component of the \texttt{retain} set, we leverage the large-scale CommonPool corpus from the DataComp benchmark~\cite{DataComp}. This corpus contains approximately 1.3 billion image-text pairs, of which we recovered around 80\%. Following the DataComp baseline strategy, we apply CLIPScore-based filtering, keeping the top 40\% of pairs. This results in a high-quality dataset of approximately 400 million pairs, which we use as the foundation for both models' general knowledge. 
With the SALMU data splits fully defined, the training sets of both models are as follows:
\begin{itemize}
    \item The \textit{\textbf{Clean}} model, which represents the ideal unlearned state, is trained exclusively on the \texttt{retain} set.
    \item The \textit{\textbf{Compromised}} model, which is polluted with sensitive information, is trained on the union of the \texttt{retain} and \texttt{sensitive} sets. This model serves as the starting point for all unlearning interventions.
\end{itemize}


\paragraph{Training Setup.} Both the \textit{Clean} and \textit{Compromised} models adopt the ViT/B-16 architecture, consistent with the default configuration used for the large-scale flavor of the DataComp benchmark~\cite{DataComp}. We follow the training setup from the OpenCLIP repository~\cite{OpenCLIPRepo} and train both models from scratch for 32 epochs, where the \textit{Clean} model uses \texttt{retain} (400M) and the \textit{Compromised} model uses \texttt{retain} (400M) + \texttt{sensitive} (60K). The training process takes approximately 65 hours using 128 NVIDIA H100 GPUs. Key training hyperparameters are as follows: \texttt{batch\_size}: 8192; \texttt{learning\_rate}: 5e-4; and \texttt{warmup\_steps}: 500.

Most of these settings match those used in the large-scale DataComp ViT/B-16 training regime, except for the number of epochs that we set to 32 as in the original CLIP model~\cite{OpenAICLIPOriginal} and OpenCLIP~\cite{OpenCLIPRepo} when trained on similar data scales (400M), ensuring both fidelity to real-world CLIP pretraining and comparability with existing CLIP model baselines.

\section{Data Validation and Domain Gap Analysis}
\label{sec:data_validation}

In Section \ref{subsec:model-data-validation} of the main paper, we discuss the validity of using synthetic faces as a proxy for real portraits. To quantify the domain gap, we compared the cosine similarity distributions of 100 real portraits from the FHIBE dataset~\cite{FHIBE} against 100 randomly sampled images from our synthetic \texttt{sensitive} set. Both sets were paired with the following generic captions to isolate visual domain differences from semantic association effects:
\begin{itemize}
    \item \textit{``person''}
    \item \textit{``a person''}
    \item \textit{``real person''}
    \item \textit{``a photo of a person''}
    \item \textit{``an image of a person''}
\end{itemize}

We performed a two-sample Kolmogorov-Smirnov (KS) test to compare the distributions generated by both the \textit{Clean} and \textit{Compromised} models. The results are as follows:

\paragraph{Clean Model Analysis.}
As visually demonstrated in Figure \ref{fig:domain-gap} of the main paper, the distributions for the \textit{Clean} model overlap significantly. Quantitatively, the test yields a KS statistic of $0.0900$ with a $p$-value of $0.8154$. Since $p > 0.05$, we cannot reject the null hypothesis, indicating no statistically significant difference between the embedding distributions of real and synthetic faces.
\begin{itemize}
    \item Real Images: $\mu=0.2128, \sigma=0.0175$
    \item Synthetic Images: $\mu=0.2128, \sigma=0.0141$
\end{itemize}

\begin{figure}[h]
    \centering
    \includegraphics[width=\linewidth]{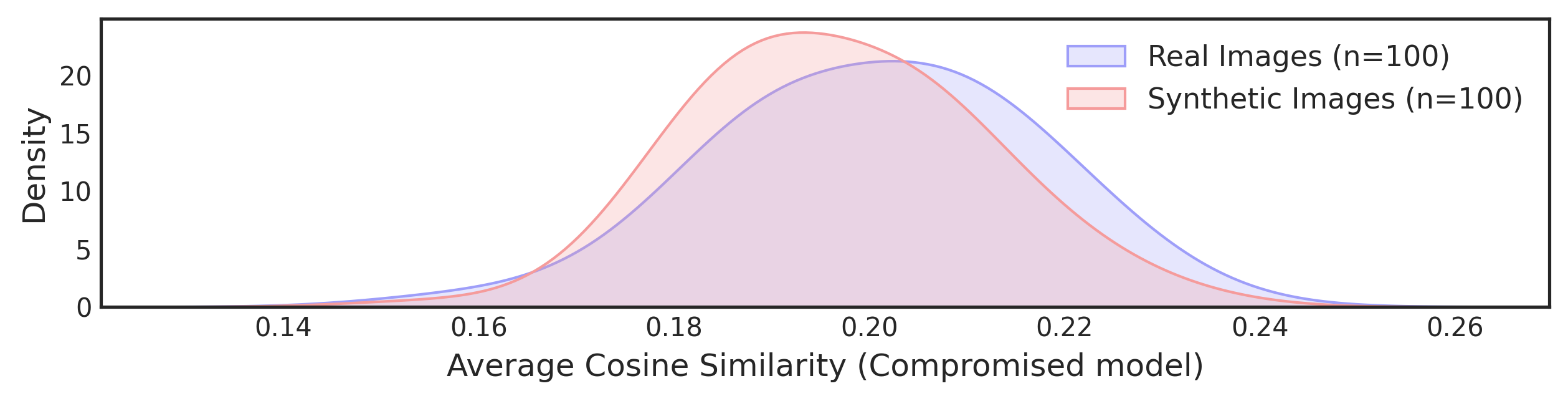} 
    \caption{Average cosine similarity of Real vs. Synthetic images with generic captions for the \textit{Compromised} model.}
    \label{fig:compromised_domain_gap}
\end{figure}

\paragraph{Compromised Model Analysis.}
We repeated this analysis using the \textit{Compromised} model to ensure that the inclusion of sensitive data during training did not distort the visual domain representation. As shown in Figure~\ref{fig:compromised_domain_gap}, the distributions remain aligned. The test yielded a KS statistic of $0.1700$ with a $p$-value of $0.1112$. Again, with $p > 0.05$, the distributions are statistically indistinguishable.
\begin{itemize}
    \item Real Images: $\mu=0.2005, \sigma=0.0154$
    \item Synthetic Images: $\mu=0.1966, \sigma=0.0139$
\end{itemize}

These results confirm that our synthetic dataset serves as a valid, privacy-safe proxy for real-world portraits in the context of embedding-based unlearning.

\subsection{Clean Model Stability}
\label{subsec:clean_stability}

To assess whether the \textit{Clean} model's efficacy metrics reflect a stable baseline rather than a single-run artifact, we compared it against an independent OpenCLIP ViT-B/16 model trained on LAION-400M. As shown in Table~\ref{tab:clean_stability}, this independent model's sensitive-set metrics closely track our \textit{Clean} model (e.g., RetFail 0.004 vs.\ 0.001), confirming that the evaluation targets represent a robust ``clean regime.'' While strict statistical indistinguishability from a retrained model is the theoretical ideal, these results indicate that the practical goal is restoring the model to this stable baseline state.

\begin{table}[h]
\centering
\small
\setlength{\tabcolsep}{3pt}
\begin{tabular}{lccccc}
\toprule
\textbf{Model} & \textbf{RetFail} & \textbf{AssocStr} & \textbf{ACS} & \textbf{IdZSC} & \textbf{CoreAssoc} \\
\midrule
\textit{Clean} & \textbf{0.001} & 0.142 & 0.602 & 0.011 & 0.151 \\
LAION-400M & 0.004 & 0.149 & 0.632 & 0.005 & 0.169 \\
\textit{Compromised} & 0.236 & 0.322 & 0.951 & 0.084 & 0.183 \\
\bottomrule
\end{tabular}
\caption{Sensitive-set efficacy metrics for the \textit{Clean}, \textit{Compromised}, and an independent OpenCLIP (LAION-400M) model. The similarity between \textit{Clean} and LAION-400M confirms the target scores are stable and representative.}
\label{tab:clean_stability}
\end{table}

\section{Statistical Analysis Framework}
\label{subsec:stat_framework}

To rigorously interpret the benchmark results, we define a formal statistical framework for comparing any \textit{Unlearned} model against the foundational \textit{Compromised} and \textit{Clean} models. The ultimate goal is for an \textit{Unlearned} model to become statistically indistinguishable from its ideal target: the \textit{Clean} model for efficacy metrics and the original \textit{Compromised} model for utility metrics. Consequently, the ideal outcome for all statistical tests is a non-significant result (p-value $\geq$ 0.05), indicating that the unlearning method has perfectly reached its target state for a given metric.

\paragraph{Analyzing Forgetting Efficacy.} The goal is for an \textit{Unlearned} model's behavior on sensitive data to converge to that of the \textit{Clean} model.
\begin{itemize}
    \item For the retrieval metric (RetFail), we perform a two-sided Wilcoxon signed-rank test on the paired sample-wise ranks. We test if the distribution of ranks for the \textit{Unlearned} model is statistically different from that of the \textit{Clean} model. The ideal result is non-significant, as this indicates that the \textit{Unlearned} model’s retrieval performance has successfully converged to the state of the \textit{Clean} model, which never saw the sensitive data.
    \item For metrics based on boolean outcomes (ACS, IdZSC), we compare the sample-wise results using McNemar's test to determine if there is a significant difference in outcome rates. The ideal result is non-significant, indicating the models are statistically equivalent.
    \item For metrics based on similarity scores (AssocStr, CoreAssoc), we compare the score distributions using a two-sample Kolmogorov-Smirnov (K-S) test. The ideal result is non-significant, suggesting the distributions are identical.
\end{itemize}

\paragraph{Analyzing Utility Impact.} The goal is to preserve the general capabilities of the \textit{Compromised} model, ensuring that the unlearning process does not introduce any significant collateral damage or unintended changes.
\begin{itemize}
    \item To assess for any change in general knowledge (GenKnow), we use McNemar's test to compare the prediction outcomes of the \textit{Unlearned} model against the \textit{Compromised} model on ImageNet-1K. The ideal result is non-significant.
    \item For the remaining utility metrics (InterIdSim, IntraIdSim, VisIdInt, FragSim), we assess for any performance change using a two-sided paired t-test. This test determines if there is any significant difference (either an increase or a decrease) in the mean of the sample-wise similarity scores between the \textit{Unlearned} and \textit{Compromised} models. The ideal result remains non-significant, indicating that the model's performance on these utility tasks has been perfectly preserved.
\end{itemize}

\section{Baselines and Methods for CLIP Unlearning}
\label{sec:baselines}

We evaluate several unlearning strategies and a finetuning baseline on SALMUBench. Since established multimodal unlearning frameworks (e.g., MultiDelete~\cite{cheng2024multidelete}) were not originally designed for CLIP's specific architecture, we performed the necessary adaptations to map their objectives to the dual-encoder setting. We evaluate each method under varying computational budgets (1$\times$, 5$\times$, 10$\times$, 15$\times$, 20$\times$).

\paragraph{Generic Captions.}
This baseline fine-tunes the model on \texttt{forget} images paired with generic, non-sensitive captions precomputed with BLIP~\cite{BLIP-image-captioning-large}. The training objective is to minimize the standard contrastive loss, using AdamW optimizer with a learning rate of \texttt{3e-7} and a batch size of 256.

\paragraph{Shuffled Captions.}
This method trains on batches from the \texttt{forget} set where captions are randomly shuffled, creating hard-negative pairs to break the learned associations. The model minimizes the contrastive loss on these shuffled pairs using the AdamW optimizer, a batch size of 256, and a learning rate of \texttt{5e-7}.

\paragraph{Direct Similarity Minimization.}
This method’s objective is to directly minimize the mean cosine similarity between the sensitive image and text embeddings. This provides a direct ``anti-similarity'' signal trained with the AdamW optimizer, a batch size of 256, and a learning rate of \texttt{3e-8}.

\paragraph{Negative Gradient}. Inspired by the approach presented in~\cite{golatkar2020eternal}, it unlearns by reversing the gradient direction on the \texttt{forget} set. During training, we apply gradient ascent (maximize the loss instead of minimizing it) to push the model away from the representations learned on the \texttt{forget} data. We use batch size of 256 and AdamW optimizer with a learning rate of 3e-8.

\paragraph{Finetuning}~\cite{golatkar2020eternal}.
It finetunes the pretrained model on an 80K-example retain set (randomly sampled from the original training data, excluding the forget set), using a low learning rate of 5e-7, weight decay of 0.001, and batch size of 32.

\paragraph{Descent to Delete}~\cite{neel2021descent}.
This is a weight-space unlearning method that perturbs model parameters with carefully scaled Gaussian noise to obscure the influence of specific data. The noise level is computed based on model smoothness, curvature, and the size of the original training set to ensure that the final model is statistically indistinguishable from one trained without the forgotten data.

\paragraph{VLUnlearn}~\cite{cheng2024multidelete}.
It enforces forgetting by increasing the distance between image-text embeddings for the \texttt{forget} set. The original method relies on visual-language models with cross-attention fusion layers (e.g., ALBEF or BLIP) to compute joint representations. Since CLIP lacks such fusion modules, we modify the implementation to use the average of image and text features as a proxy for the fused encoder output. We maintain the original training objectives: modality decoupling, multimodal knowledge retention, and unimodal knowledge retention. We use the default hyper-parameters provided by the authors in the official repository\footnote{\url{https://github.com/CLU-UML/MultiDelete}}.

\paragraph{DELETE} ~\cite{zhou2025decoupled}.
DELETE is an unimodal class-centric unlearning method that decouples the forgetting and retention objectives. In our CLIP adaptation, the \texttt{forget} set is used to minimize the image-text similarity for sensitive pairs (forgetting), while the \texttt{retain} set is distilled from the frozen original model to preserve behavior on non-sensitive pairs (retention). We omit the original logit-masking step and instead operate directly on CLIP’s cosine similarity scores. The model is trained with MSE losses on both forgetting and retention terms, using the AdamW optimizer with a learning rate of \texttt{3e-8} and batch size of 256.

\paragraph{CLIPErase}~\cite{yang-etal-2025-cliperase}.
A targeted unlearning framework for contrastive multimodal models. It jointly optimizes three objectives: (1) reducing similarity for sensitive image-text pairs (forgetting), (2) matching the original model’s similarity on the retain set (retention), and (3) preserving unimodal image and text embeddings to prevent representation drift (consistency). In our implementation, we compute all three terms using MSE losses on CLIP’s cosine similarity scores and embeddings, following the paper’s design but without modifying the model architecture. Training uses the AdamW optimizer with a learning rate of \texttt{3e-8} and batch size of 256.

\begin{table*}[ht]
\centering
\small
\renewcommand{\arraystretch}{1}
\setlength{\tabcolsep}{2.5pt} 
\begin{tabular}{lr >{\columncolor{gray!20}}ccccc >{\columncolor{gray!20}}ccccc}
\toprule
& & \multicolumn{5}{c}{\textbf{Forgetting Efficacy}} & \multicolumn{5}{c}{\textbf{Utility Impact}} \\
\cmidrule(lr){3-7} \cmidrule(lr){8-12}
\textbf{Method} & \textbf{Budget} & \textbf{RetFail} & \textbf{AssocStr} & \textbf{ACS} & \textbf{IdZSC} & \textbf{CoreAssoc} & \textbf{GenKnow} & \textbf{InterIdSim} & \textbf{IntraIdSim} & \textbf{VisIdInt} & \textbf{FragSim} \\
\midrule
\textit{\textbf{Target Scores}} & \textbf{-} & \textbf{0.001} & \textbf{0.142} & \textbf{0.602} & \textbf{0.011} & \textbf{0.151} & \textbf{0.638} & \textbf{0.321} & \textbf{0.321} & \textbf{0.305} & \textbf{0.280} \\
\midrule
\multirow{5}{5em}{Generic Captions} & 1$\times$ & 0.143 & 0.308 & 0.932 & 0.073 & 0.179 & 0.635 & 0.308 & 0.308 & 0.325 & 0.292 \\
& 5$\times$ & 0.055 & 0.288 & 0.907 & 0.056 & 0.185 & 0.629 & 0.288 & 0.288 & 0.338 & 0.310 \\
& 10$\times$ & 0.042 & 0.275 & 0.896 & 0.050 & 0.185 & 0.625 & 0.276 & 0.275 & 0.341 & 0.313 \\
& 15$\times$ & 0.038 & 0.267 & 0.885 & 0.044 & 0.184 & 0.621 & 0.269 & 0.267 & 0.342 & 0.315 \\
& 20$\times$ & 0.035 & 0.258 & 0.877 & 0.042 & 0.182 & 0.617 & 0.260 & 0.258 & 0.342 & 0.313 \\
\midrule
\multirow{5}{5em}{Shuffled Captions} & 1$\times$ & 0.005 & 0.213 & 0.644 & 0.003 & 0.192 & 0.587 & 0.213 & 0.213 & 0.210 & 0.206 \\
& 5$\times$ & 0.004 & 0.212 & \textbf{0.593} & 0.002 & 0.200 & 0.548 & 0.212 & 0.212 & 0.206 & 0.203 \\
& 10$\times$ & 0.005 & 0.217 & 0.574 & 0.002 & 0.206 & 0.513 & 0.217 & 0.217 & 0.207 & 0.204 \\
& 15$\times$ & 0.006 & 0.229 & 0.564 & 0.001 & 0.216 & 0.479 & 0.229 & 0.229 & 0.215 & 0.213 \\
& 20$\times$ & 0.013 & 0.250 & 0.551 & 0.000 & 0.231 & 0.432 & 0.250 & 0.250 & 0.229 & 0.227 \\
\midrule
\multirow{5}{5em}{Direct\\Similarity Minimization} & 1$\times$ & 0.087 & 0.262 & 0.948 & 0.074 & 0.167 & \textbf{0.638} & 0.264 & 0.263 & 0.283 & 0.259 \\
& 5$\times$ & \textbf{0.001} & -0.429 & \textbf{0.593} & 0.002 & 0.012 & 0.615 & -0.420 & -0.425 & -0.038 & -0.031 \\
& 10$\times$ & \textbf{0.001} & -0.860 & 0.529 & 0.000 & -0.061 & 0.573 & -0.855 & -0.858 & -0.263 & -0.250 \\
& 15$\times$& \textbf{0.001} & -0.944 & 0.520 & 0.000 & -0.098 & 0.526 & -0.942 & -0.943 & -0.323 & -0.310 \\
& 20$\times$ & \textbf{0.001} & -0.967 & 0.516 & 0.000 & -0.140 & 0.469 & -0.966 & -0.967 & -0.355 & -0.344 \\
\midrule
\multirow{5}{5em}{Negative Gradient} & 1$\times$ & 0.179 & 0.313 & 0.929 & 0.072 & 0.189 & \textbf{0.638} & \textbf{0.314} & \textbf{0.313} & \textbf{0.306} & \textbf{0.277} \\
& 5$\times$ & 0.009 & 0.055 & 0.657 & 0.023 & 0.204 & 0.630 & 0.063 & 0.061 & 0.194 & 0.171 \\
& 10$\times$ & 0.007 & -0.206 & 0.554 & 0.014 & 0.164 & 0.619 & -0.201 & -0.196 & 0.078 & 0.082 \\
& 15$\times$ & 0.006 & -0.450 & 0.526 & \textbf{0.009} & 0.143 & 0.604 & -0.449 & -0.440 & -0.049 & -0.024 \\
& 20$\times$ & 0.006 & -0.646 & 0.511 & 0.006 & \textbf{0.149} & 0.587 & -0.649 & -0.638 & -0.162 & -0.128 \\
\midrule
\multirow{5}{5em}{Finetuning} & 1$\times$ & 0.004 & 0.218 & 0.647 & 0.006 & 0.222 & 0.646 & 0.220 & 0.220 & 0.287 & 0.266 \\
& 5$\times$ & 0.003 & 0.208 & 0.646 & 0.005 & 0.212 & \textbf{0.638} & 0.209 & 0.209 & 0.292 & 0.271 \\
& 10$\times$ & 0.002 & 0.203 & 0.647 & 0.005 & 0.209 & 0.637 & 0.204 & 0.204 & 0.294 & 0.273 \\
& 15$\times$ & 0.002 & 0.200 & 0.647 & 0.005 & 0.208 & 0.637 & 0.201 & 0.201 & 0.295 & 0.274 \\
& 20$\times$ & 0.002 & \textbf{0.199} & 0.648 & 0.004 & 0.207 & 0.637 & 0.201 & 0.201 & 0.295 & 0.274 \\
\midrule
Descent to Delete & - & 0.008 & 0.227 & 0.644 & 0.007 & 0.230 & 0.644 & 0.228 & 0.228 & 0.277 & 0.258 \\
\midrule
\multirow{4}{4em}{VLUnlearn} & 5$\times$ & \textbf{0.001} & 0.210 & 0.542 & 0.006 & 0.211 & 0.638 & 0.210 & 0.210 & 0.260 & 0.248 \\
& 10$\times$ & \textbf{0.001} & 0.209 & 0.528 & 0.007 & 0.212 & 0.641 & 0.209 & 0.209 & 0.262 & 0.248 \\
& 15$\times$ & \textbf{0.001} & 0.209 & 0.524 & 0.007 & 0.213 & 0.641 & 0.208 & 0.209 & 0.262 & 0.248 \\
& 20$\times$ & \textbf{0.001} & 0.208 & 0.525 & 0.006 & 0.212 & 0.643 & 0.208 & 0.208 & 0.263 & 0.248 \\
\midrule
\multirow{4}{4em}{DELETE} & 5$\times$ & \textbf{0.001} & 0.022 & 0.547 & 0.000 & 0.047 & 0.632 & 0.023 & 0.023 & 0.221 & 0.217 \\
& 10$\times$ & \textbf{0.001} & 0.014 & 0.536 & 0.000 & 0.044 & 0.642 & 0.015 & 0.014 & 0.240 & 0.234 \\
& 15$\times$ & \textbf{0.001} & 0.013 & 0.529 & 0.000 & 0.049 & 0.643 & 0.014 & 0.013 & 0.249 & 0.242 \\
& 20$\times$ & \textbf{0.001} & 0.012 & 0.524 & 0.000 & 0.050 & 0.643 & 0.013 & 0.013 & 0.248 & 0.240 \\
\midrule
\multirow{4}{4em}{CLIPErase}& 5$\times$ & \textbf{0.001} & 0.023 & 0.550 & 0.000 & 0.051 & 0.634 & 0.024 & 0.024 & 0.225 & 0.222 \\
& 10$\times$ & \textbf{0.001} & 0.019 & 0.534 & 0.000 & 0.056 & 0.642 & 0.020 & 0.020 & 0.245 & 0.237 \\
& 15$\times$ & \textbf{0.001} & 0.018 & 0.527 & 0.000 & 0.065 & 0.643 & 0.020 & 0.019 & 0.253 & 0.243 \\
& 20$\times$ & \textbf{0.001} & 0.019 & 0.525 & 0.000 & 0.069 & 0.642 & 0.020 & 0.020 & 0.253 & 0.243 \\
\bottomrule
\end{tabular}
\caption{Performance of baseline unlearning methods across different computational budgets.}
\label{tab:unlearning_performance}
\end{table*}

\begin{table*}[t]
\centering
\small
\setlength{\tabcolsep}{2.5pt} 
\begin{tabular}{lr >{\columncolor{gray!20}}ccccc >{\columncolor{gray!20}}ccccc}
\toprule
& & \multicolumn{5}{c}{\textbf{Forgetting Efficacy}} & \multicolumn{5}{c}{\textbf{Utility Impact}} \\
\cmidrule(lr){3-7} \cmidrule(lr){8-12}
\textbf{Method} & \textbf{Budget} & \textbf{RetFail} & \textbf{AssocStr} & \textbf{ACS} & \textbf{IdZSC} & \textbf{CoreAssoc} & \textbf{GenKnow} & \textbf{InterIdSim} & \textbf{IntraIdSim} & \textbf{VisIdInt} & \textbf{FragSim} \\
\midrule
\textit{\textbf{Clean (Reference)}} & \textbf{-} & \textbf{0.001} & \textbf{0.142} & \textbf{0.602} & \textbf{0.011} & \textbf{0.151} & \textbf{0.633} & \textbf{0.143} & \textbf{0.143} & \textbf{0.331} & \textbf{0.309} \\
\midrule
Generic Captions & 5$\times$ & \textbf{0.001} & \textbf{0.147} & \textbf{0.616} & \textbf{0.011} & \textbf{0.156} & 0.626 & \textbf{0.148} & \textbf{0.148} & \textbf{0.324} & \textbf{0.301} \\
Shuffled Captions & 5$\times$ & 0.002 & 0.185 & 0.521 & 0.001 & 0.185 & 0.606 & 0.185 & 0.185 & 0.183 & 0.184 \\
Direct Sim. Min. & 5$\times$ & \textbf{0.001} & -0.371 & 0.512 & 0.005 & -0.259 & 0.611 & -0.369 & -0.368 & 0.074 & 0.074 \\
Negative Gradient & 5$\times$ & \textbf{0.001} & -0.114 & 0.529 & 0.007 & -0.045 & \textbf{0.629} & -0.116 & -0.112 & 0.254 & 0.248 \\
\bottomrule
\end{tabular}

\caption{Performance of 4 baseline unlearning methods applied to the \textit{Clean} model (5$\times$ budget). The first row lists the metrics of the unmodified \textit{Clean} model. Deviations from these values indicate unintended damage to the model's embedding space.}
\label{tab:clean_5x_unlearning_performance}
\end{table*}

\section{Experimental Results}

\subsection{Unlearning Baselines Performance}

To complement the main paper's evaluation, where we focus on the $5\times$ compute budget due to its generally favorable trade-off between forgetting efficacy and utility preservation, we provide in Table~\ref{tab:unlearning_performance} the complete performance breakdown across all considered computational budgets ($1\times$ to $20\times$). This extended view offers insights into how each method scales under increasing resource availability. 

In this analysis, the `Target Scores' serve as the ideal empirical bounds: they correspond to the \textit{Clean} model for Forgetting Efficacy metrics (representing perfect removal) and the \textit{Compromised} model for Utility Impact metrics (representing original knowledge preservation).

While some methods exhibit relatively stable behavior across budgets (e.g., Finetuning, DELETE, CLIPErase), others show strong budget-dependent dynamics. Notably, Direct Similarity Minimization performs best at low budgets but deteriorates significantly as training progresses -- highlighting the method’s tendency to over-optimize, resulting in excessive structural damage to the embedding space.

Similarly, methods like Generic Captions and Shuffled Captions show mild gains with larger budgets, though their forgetting performance often saturates early, suggesting limited capacity for deeper erasure without stronger supervisory signals.

Overall, we chose the $5\times$ budget for the main paper as it tends to represent the best trade-off across most methods: high forgetting performance without significant collateral damage. However, we emphasize that each method can, in principle, be re-tuned per budget setting. For example, hyperparameters like learning rate, number of epochs, or regularization strength could be adjusted to yield more optimal trade-offs at higher budgets. Nevertheless, this study intentionally fixes per-method hyperparameters across all budgets to isolate the effect of compute scale and avoid confounding factors. We leave hyperparameter tuning for budget-aware unlearning as a direction for future SALMUBench submissions.

\subsection{Sanity Check: Unlearning on the Clean Model}
\label{subsec:sanity_check}

A critical failure mode identified in our main paper is the tendency of unlearning algorithms to damage model utility regardless of whether the sensitive data was actually present. We applied unlearning baselines to the \textit{Clean} model. Since this model represents the desired unlearned state, a safe method should effectively act as an identity function, preserving the metrics of the original \textit{Clean} model.

Table~\ref{tab:clean_5x_unlearning_performance} presents the results. The first row displays the original scores of the \textit{Clean} model, serving as the reference for stability.

\paragraph{Analysis.}
The results reveal a clear dichotomy between methods, with a notable nuance regarding utility preservation:

\begin{enumerate}
    \item \textbf{Low-Risk Methods:} Generic Captions proves to be the most stable approach regarding embedding structure, maintaining scores very close to the \textit{Clean} model reference (e.g., InterIdSim $0.148$ vs. Reference $0.143$). However, it is not entirely cost-free, as it induces a slight degradation in general knowledge (GenKnow drops from $0.633$ to $0.626$). Shuffled Captions causes more significant drift across both utility and structural metrics, confirming that the noise from mismatched pairs is more harmful than the use of generic labels.
    
    \item \textbf{Catastrophically Damaging Methods:} Direct Similarity Minimization and Negative Gradient fail this check significantly. Negative Gradient presents a deceptive profile: while it preserves GenKnow reasonably well ($0.629$), it aggressively distorts the embedding space, driving several metrics to negative values. Direct Similarity Minimization is even more destructive, causing a larger drop in utility and an even more extreme inversion of structure. The fact that these methods drive the \textit{Clean} model's natural baseline alignment (AssocStr $0.142$) into negative territory ($-0.114$ and $-0.371$) confirms they do not selectively unlearn but rather blindly disrupt the feature space.
\end{enumerate}

\section{Learning-Rate Sensitivity Ablation}
\label{sec:lr_ablation}

To validate our failure mode taxonomy beyond a single hyperparameter setting, we performed a learning-rate (LR) sweep at 5$\times$ budget for two representative methods: Generic Captions (classified as Type~3: Ineffective) and Negative Gradient (classified as Type~2: Over-generalized). For each method, we varied the learning rate across a wide range and measured RetFail (forgetting efficacy) and GenKnow (utility preservation).

\begin{figure}[h]
    \centering
    \includegraphics[width=\linewidth]{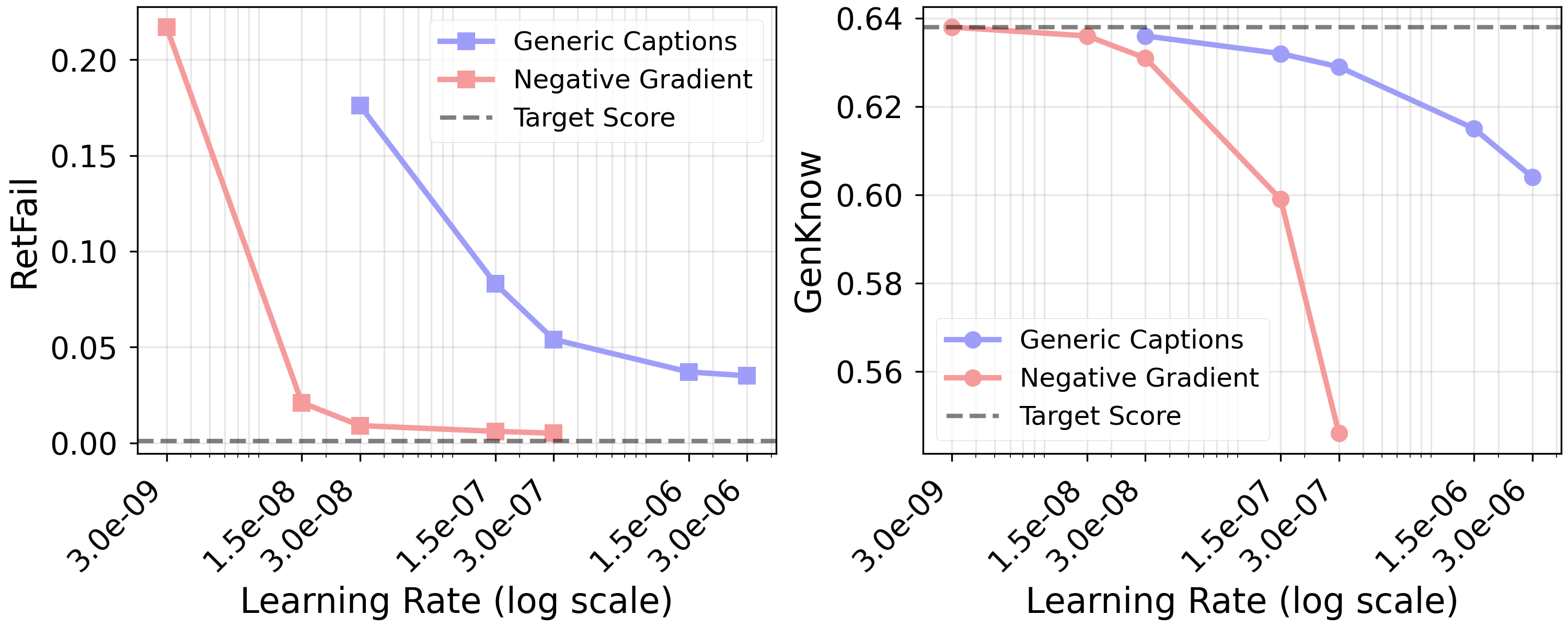}
    \caption{Learning-rate sensitivity (5$\times$ budget): RetFail (left) and GenKnow (right) for Generic Captions and Negative Gradient. Dashed lines indicate target scores. Generic Captions hits a forgetting ceiling regardless of LR, confirming Type~3 (Ineffective). Negative Gradient shows that reaching the forgetting target forces GenKnow well below the target, confirming Type~2 (Over-generalized).}
    \label{fig:lr_ablation}
\end{figure}

As shown in Figure~\ref{fig:lr_ablation}, the results confirm our taxonomy. Generic Captions exhibits a clear performance ceiling: increasing the learning rate degrades utility (GenKnow) without ever reaching the forgetting target (RetFail), confirming its classification as Type~3 (Ineffective). Negative Gradient confirms Type~2 (Over-generalized): while intermediate learning rates might appear balanced, reaching the target RetFail forces AssocStr well below the \textit{Clean} baseline and eventually collapses GenKnow. These results demonstrate that the identified failure modes are robust properties of the methods, not artifacts of a single hyperparameter choice.

\end{document}